\documentclass[letterpaper]{article} % DO NOT CHANGE THIS
\usepackage{aaai24}  % DO NOT CHANGE THIS
\usepackage{times}  % DO NOT CHANGE THIS
\usepackage{helvet}  % DO NOT CHANGE THIS
\usepackage{courier}  % DO NOT CHANGE THIS
\usepackage[hyphens]{url}  % DO NOT CHANGE THIS
\usepackage{graphicx} % DO NOT CHANGE THIS
\usepackage{natbib}  % DO NOT CHANGE THIS AND DO NOT ADD ANY OPTIONS TO IT
\usepackage{caption} % DO NOT CHANGE THIS AND DO NOT ADD ANY OPTIONS TO IT
\usepackage{algorithm}
\usepackage{algorithmic}

\usepackage{booktabs}       % professional-quality tables
\usepackage{amsfonts}       % blackboard math symbols
\usepackage{nicefrac}       % compact symbols for 1/2, etc.
\usepackage{microtype}      % microtypography
\usepackage{xcolor}         % colors
\usepackage{adjustbox}
\usepackage{bm}
\usepackage{graphicx}
\usepackage{amsmath}
\usepackage{amsthm}

\usepackage{subfigure}
\usepackage{makecell}
\usepackage{multirow}
\usepackage{amssymb}

\usepackage{newfloat}
\usepackage{listings}
\DeclareCaptionStyle{ruled}{labelfont=normalfont,labelsep=colon,strut=off} % DO NOT CHANGE THIS
\floatstyle{ruled}
\newfloat{listing}{tb}{lst}{}
\floatname{listing}{Listing}
\newcommand{\ie}{\textit{i.e.}}
\newcommand{\eg}{\textit{e.g.}}
\newcommand{\red}{\textcolor{red}}
\title{LimeAttack: Local Explainable Method for  Textual  Hard-Label Adversarial Attack}
\author{
    Hai Zhu\textsuperscript{\rm 1 \rm 3}\thanks{Corresponding author.},
    Qingyang Zhao\textsuperscript{\rm 2},
    Weiwei Shang\textsuperscript{\rm 1},
    Yuren Wu\textsuperscript{\rm 3},
    Kai Liu\textsuperscript{\rm 4}
}
\affiliations{
    \textsuperscript{\rm 1} University of Science and Technology of China\\
    \textsuperscript{\rm 2} Xidian University\\
    \textsuperscript{\rm 3}Ping An Technology\\
    \textsuperscript{\rm 4}Lazada\\
    SA21218029@mail.ustc.edu.cn,21151213588@stu.xidian.edu.cn,wwshang@ustc.edu.cn,leifive@163.com,baiyang.lk@alibaba-inc.com
}

\usepackage{bibentry}
\begin{document}

\maketitle
\begin{abstract}
    Natural language processing models are vulnerable to adversarial examples. Previous textual adversarial attacks adopt model internal information 
 (gradients or  confidence scores) to generate adversarial examples. However, this information is unavailable in the real world. Therefore, we focus on a more realistic and challenging setting, named hard-label attack, in which the attacker can only query the model and obtain a discrete prediction label. Existing hard-label attack algorithms tend to initialize adversarial examples by random substitution and then utilize complex heuristic algorithms to optimize the adversarial perturbation. These methods require a lot of  model queries and the attack success rate is restricted by adversary initialization. In this paper,  we propose a novel hard-label attack algorithm named LimeAttack, which leverages a local explainable method to approximate word importance ranking, and then adopts beam search to find the optimal solution. Extensive experiments show that LimeAttack achieves the better attacking performance compared with existing hard-label attack under the same query budget. In addition, we evaluate the effectiveness of LimeAttack on large language models and some defense methods, and results indicate that adversarial examples remain a significant threat to  large language models. The adversarial examples crafted by LimeAttack  are highly transferable and  effectively improve model robustness in adversarial training. %Code are provided in \url{https://anonymous.4open.science/r/limeattack-FE2C}
\end{abstract}
\section{Introduction}
Deep Neural Networks (DNNs) are widely applied in the  natural language processing field and have achieved great success~\cite{kim-2014-convolutional,bert2019,minaee2021deep,6795963}. However, DNNs are vulnerable to adversarial examples, which are correctly classified samples altered by some slight perturbations~\cite{2020-textfooler,papernot2017practical,kurakin2016adversarial}. These adversarial perturbations are imperceptible to humans but can mislead the model. Adversarial examples seriously threaten the  robustness and reliability of DNNs, especially in some security-critical applications (\eg, autonomous driving and toxic text detection ~\cite{yang2021besa,kurakin2018adversarial}).  Therefore, adversarial examples have attracted enormous attention on adversarial attacks and defenses in computer vision, natural language processing and speech~\cite{szegedy2013intriguing,carlini2018audio,texthackeryu2022learning}. It is more challenging to  craft textual adversarial examples due to the discrete nature of language along with the presence of lexical, semantic, and fluency constraints.

According to different scenarios, textual adversarial attacks can be briefly  divided into white-box attacks, score-based attacks and hard-label attacks. In a white-box setting, the attacker utilizes the model's parameters and  gradients to generate adversarial examples~\cite{goodman2020fastwordbug,jiang2020smart}. Score-based attacks only adopt class probabilities or confidence scores to craft adversarial examples~\cite{2020-textfooler,li-etal-2020-bert-attack,2020pwds,zhu2023beamattack}. However, these attack methods perform poorly in reality due to   DNNs being deployed through application programming interfaces (APIs), and the attacker having no access to the model's parameters, gradients or probability distributions of all labels~\cite{ye2022texthoaxer}.  In contrast,  under a hard-label  scenario, the   model's internal structures, gradients, training data and even confidence scores are unavailable. The attacker can only query the black-box victim model and get a discrete prediction label, which is more  challenging and realistic.
Additionally, most realistic models (\eg, HuggingFace API, OpenAI API) usually have a limit on the number of calls. In reality, the adversarial examples attack setting is hard-label with tiny model queries.

Some hard-label attack algorithms have been proposed~\cite{texthackeryu2022learning,ye2022texthoaxer,hlbbmaheshwary2021generating,ye2022leapattack}. They follow two-stages strategies: i) generate  low-quality adversarial examples by randomly replacing several original words with synonyms, and then ii) adopt complex heuristic algorithms (\eg, genetic algorithm) to optimize the adversary perturbation. Therefore, these attack methods usually require a lot of queries and the attack success rate and quality of adversarial examples are limited by adversary initialization. On the contrary, score-based attacks calculate the word importance based on  the change in confidence scores after deleting one word. Word importance ranking  improves attack efficiency by preferring to  attack words that have a significant impact on  the  model's predictions~\cite{2020-textfooler}. However, score-based attacks cannot calculate the word importance   in a hard-label setting because deleting one token hardly changes the discrete prediction label. Therefore,  we want to investigate such a problem: \emph{how to calculate word importance ranking in a hard-label setting to improve attack efficiency?} 

Actually, word importance ranking can reveal the decision boundary to determine the better attack path, but existing hard-label algorithms  ignore this useful information because  it is hard to obtain. Inspired by local  explainable methods~\cite{limeribeiro2016should,shaplundberg2017unified,deepliftshrikumar2016not} for DNNs, which  are often used to explain the outputs of black-box models,  aim to estimate the token sensitivity  on the benign sample. Previous study~\cite{addchai2023additive} has tried to simply replace deletion-based method with local  explainable method to calculate word importance in score-based attack. However, In Appendix B,  we have verified through experiments that local  explainable method does not have a significant advantage over deletion-based method in a score-based scenario.  Because  the probability distribution of the model's output is available, the influence of each word on the output can be well reflected by deletion-based method. Therefore, compared with score-based attacks, we think local  explainable method can play a greater advantage in hard-label attacks where deletion-based method is useless.  We adopt  the most fundamental and straightforward local explainable method, namely LIME. LIME is easy to understand and more in line with the deletion-based method proposed in score-based attacks, since our goal is to bridge the gap between score-base attacks and hard-label attacks by introducing interpretability method. 
 In fact, local  explainable methods are model-agnostic and  suitable for conducting word importance estimation for  hard-label attacks. However, there are the following difficulties in applying LIME to hard-label attacks: 1) How to allocate LIME and search queries under tiny query budget to achieve optimal results. 2) How to establish a mapping relationship between LIME and word importance in adversarial samples
without model's logits output. 3) How to sample reasonably during perturbation execution to achieve optimal results. In subsequent sessions we will explain in detail how to solve these difficulties.

In this work, we propose a novel hard-label attack algorithm named LimeAttack. The application of LIME in hard-label attacks was inspired by the score-based attacks' deletion method.  We verify the effectiveness of inside-to-outside attack path in hard-label attacks, then many excellent score-based attacks may provide hard-label attacks more insight. To evaluate  the attack performance and efficiency, we compare LimeAttack with other hard-label attacks and take several score-based attacks as references for  two NLP tasks on seven common datasets. We also evaluate LimeAttack on the currently state-of-the-art large language models (\eg, ChatGPT). Experiments show that LimeAttack achieves the highest attack success rate compared to other baselines  under the tiny query budget. Our contributions are  summarized as follows:
\begin{itemize}
    \item We summarize the shortcomings of the existing hard-label attacks and  apply LIME to connect score-base attacks and hard-label attacks and verify the effectiveness of inside-to-outside attack path in hard-label attacks.
    \item Extensive experiments show that LimeAttack achieves higher attack success rate than existing hard-label attack algorithms  under tiny query budget. Meanwhile, adversarial examples crafted by LimeAttack are high quality and difficult for humans to distinguish. \footnote{Code is available in \url{https://github.com/zhuhai-ustc/limeattack}}
    \item In addition, we also conduct attacks and evaluations on the currently state-of-the-art large language models. Results indicate that adversarial  examples remain a significant threat to  large language models. We also have added attack performance on defense methods and convergence results of attack success rate and perturbation rate.
\end{itemize}

\section{Related Work}
\subsection{Hard-Label Adversarial Attacks}
In a hard-label setting, the attacker can only query the victim model and get a discrete prediction label. Therefore, hard-label setting is more practical and challenging.  Existing hard-label attacks contain two-stages strategies, \ie, adversary initialization and perturbation optimization. HLBB~\cite{hlbbmaheshwary2021generating} initializes an adversarial example and adopts  a genetic algorithm to optimize the perturbation. TextHoaxer~\cite{ye2022texthoaxer} and LeapAttack~\cite{ye2022leapattack}  utilizes semantic  similarity and perturbation rate as optimization objective to search for a better perturbation matrix in the continuous word embedding space. TextHacker~\cite{texthackeryu2022learning} adopts a  hybrid local search algorithm and  a word importance table learned from attack history to guide the local search.  These attack methods often require a lot of queries to reduce the perturbation rate, and the attack success rate and quality of adversary are limited by initialization. Therefore, in this work, we attempt to craft  an adversarial example directly from the benign sample. This approach can generate high-quality adversarial examples with fewer queries.
\subsection{Local Explainable Methods}
To improve DNN interpretability and aid decision-making, various methods for explaining DNNs have been proposed and broadly categorized as global or local explainable methods.  Global explainable methods focus on the model itself by using the overall knowledge about the model's architecture and parameters. On the contrary,  local methods fit a simple and interpretable model (\eg, decision tree) to a single input to measure the contribution of each token. In detail,   local explainable methods~\cite{shaplundberg2017unified,deepliftshrikumar2016not,addvstrumbelj2014explaining} associate all input tokens by defining a linear interpretability model and assumes that the contribution of each token in the input is additive. This is also called the additive feature attribution method.  In this paper,  local interpretable model-agnostic explanation (LIME)~\cite{limeribeiro2016should} is applied to calculate word importance, which is a  fundamental and representative local explainable  method. The intuition of LIME is to  generate many neighborhood samples by deleting some original words in the benign example.  These samples are then used to train a linear model where the number of features equals to the number of words in the benign sample. The parameters of this linear model are approximated to  the importance of each word. As LIME is model-agnostic, it is suitable for hard-label attacks.  

\subsection{Limitation of Existing Hard-Label Attack}
In order to intuitively compare the difference between LimeAttack and existing hard-label attack algorithms,  we create attack search path visualizations in Figure~\ref{fig.diff}. LimeAttack's search paths are represented by green lines, and they move from \textbf{inside to outside}. LimeAttack utilizes a local explainable method to learn word importance ranking and generates adversarial examples iteratively from benign samples. This helps LimeAttack to find the nearest decision boundary direction, and costs fewer model queries to attack keywords preferentially. In contrast, previous hard-label attack algorithms' search paths are represented by blue lines, and they move from \textbf{outside to inside}. These algorithms typically begin with a randomly initialized adversarial example and optimize perturbation by maximizing semantic similarity between the initialized example and the  benign sample, which requires a lot of model queries to achieve a low perturbation rate. Furthermore, their attack success rate and adversary quality are also limited by the adversary initialization. 
\begin{figure}[h]
\centering
\includegraphics[scale=0.6]{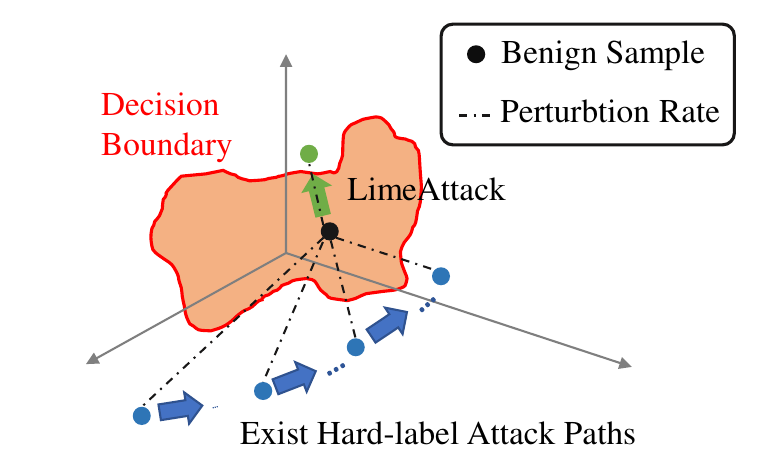} 
\caption{Search paths of existing hard-label attacks  and LimeAttack.}
\label{fig.diff}
\end{figure}
\section{Methodology}
\subsection{Problem Formulation}
Given a sentence of $n$ words $\bm{X} = [x_1,x_2,\cdots,x_n]$ and its ground truth label $Y$, an adversarial example $\bm{X'}=[x_1',x_2',\cdots,x_n']$ is crafted by replacing one or more original words with  synonyms to mislead the victim model $\mathcal{F}$. \ie,
\begin{equation}
\label{eq1.optim}
    \mathcal{F}(\bm{X'}) \neq \mathcal{F}(\bm{X}),  \quad \mathrm{s.t.} \quad D(\bm{X},\bm{X'})<\epsilon
 \end{equation}
$D(\cdot,\cdot)$ is an edit distance that measures the modifications between a benign sample $\bm{X} = [x_1,x_2,\cdots,x_n]$ and  an adversarial example $\bm{X'}=[x_1',x_2',\cdots,x_n']$:
\begin{equation}
\label{eq.pert}
    D(\bm{X},\bm{X'}) = \frac{1}{n} \sum_{i=1}^{n} \mathbb{E}(x_i,x'_i)
\end{equation}
$\mathbb{E}(\cdot,\cdot)$  is a  binary variable that equals to 0 if $x_i=x'_i$ and 1 otherwise. A high-quality adversarial example should be similar to the benign sample, and  human  readers should hardly be able to distinguish the difference. The LimeAttack belongs to the hard-label attack,  it has nothing to do with the model's parameters, gradients or confidence scores. The attacker can only query the victim model to obtain a predicted label $\hat{\bm{Y}} = \mathcal{F}(\hat{\bm{X}})$.

\subsection{The Proposed LimeAttack Algorithm}
\begin{figure*}[h]
    \centering
    \includegraphics[scale=0.5]{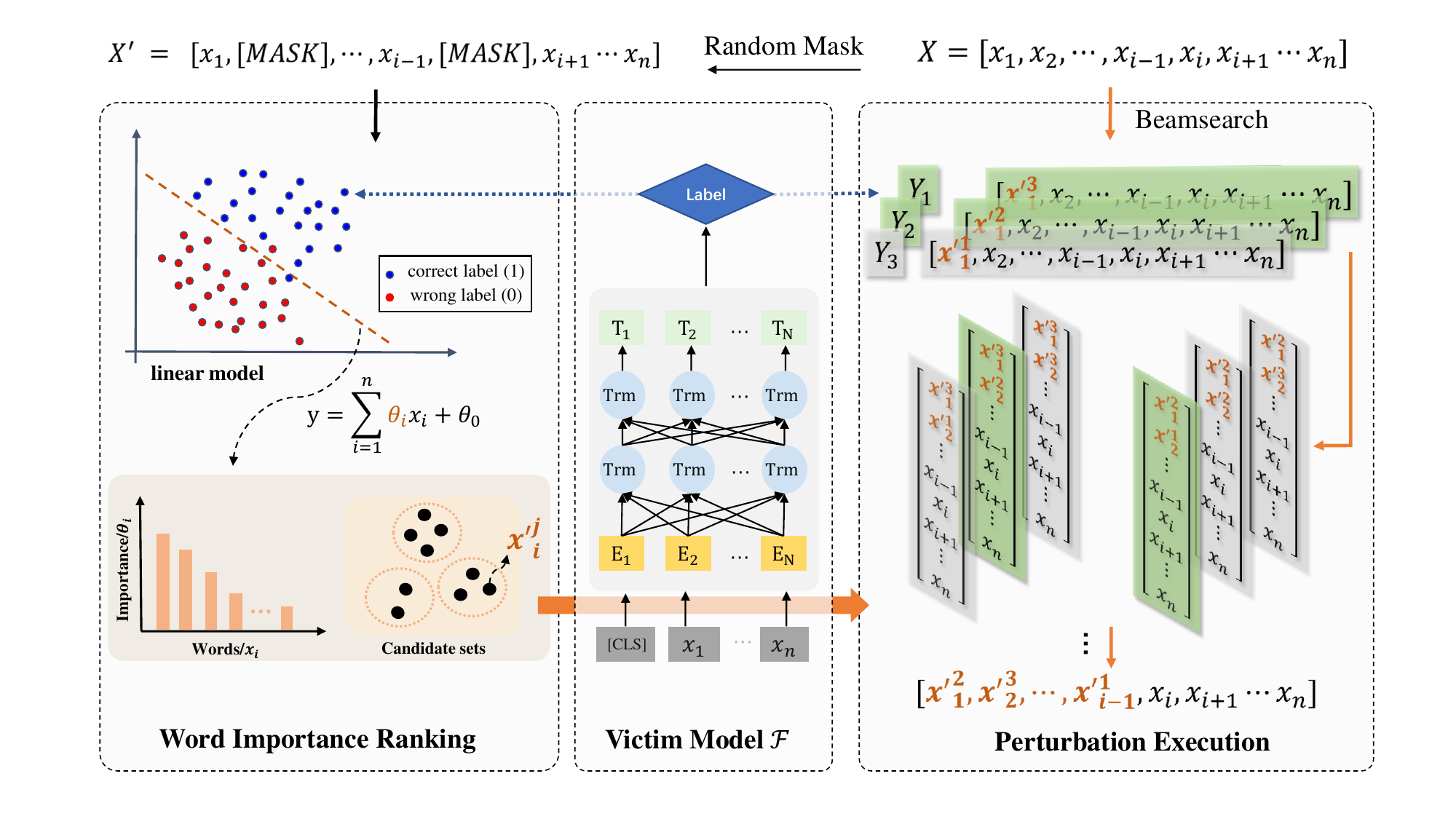}
    \caption{Overview of LimeAttack. It consists of two modules, \ie, word importance ranking and perturbation execution.  We first generate some neighborhood examples by  masking some words in the benign sample, and then adopt linear model to approximate the importance of each word $x_i$.  Then, we select candidate sets in the counter-fitted embedding space for each word. Finally, we adopt beam search (beam size $b=2$ in the figure) to generate adversarial examples iteratively.}
    \label{fig.comp}
    \end{figure*}

The overall flow chart is shown in Figure~\ref{fig.comp}. LimeAttack follows two steps, \ie,  word importance ranking and  perturbation execution. 
\subsubsection{Word Importance Ranking.}
Given a sentence of $n$ words $\bm{X}$, we assume that the contribution of all words is additive, and their sum is positively related to the model's prediction.   As shown in the Figure~\ref{fig.comp}, we generate some neighborhood samples $\mathcal{X}=[\bm{X}'_1,\bm{X}'_2,\cdots,\bm{X}'_n]$ from a benign example $\bm{X}$ by randomly replacing some words with '[MASK]'. Usually, sentences with more words often requires more neighbor samples to approximate the word importance. Therefore, we keep the number of neighborhood samples consistent with the number of tokens. We then feed $\mathcal{X}$ to the  victim model $\mathcal{F}$ to obtain  discrete prediction labels $\mathcal{\hat{Y}}=[\hat{\bm{Y}'_1},\hat{\bm{Y}'_2},\cdots,\hat{\bm{Y}'_n}]$.  Subsequently, we will  fit  a linear interpretability model to classify these neighborhood samples:
\begin{equation}
    g(\bm{X},\bm{\theta}) = \theta_0 + \sum_{i=1}^n \theta_i \mathbb{I}(x_i,\bm{X})
\end{equation}
where $\bm{\theta}$ is the parameter of the linear model, $\mathbb{I}(\cdot,\cdot)$ is a binary variable that equals to 1 if  word $x_i$ in $\bm{X}$ and 0 otherwise. Therefore, the parameter $\theta_i, i\in [1,n]$ reflects the change without word $x_i$ and  is approximated to the word importance. In Appendix O, we have verified through experiments that the linear model (such as LIME) has the same effect as some advanced interpretation methods (such as SHAP) or non-linear models  (such as decision tree) under tiny query budgets.  SHAP or non-linear models also have a higher computational complexity. The advantages of some advanced interpretation methods  or non-linear models  will only be reflected when there are a large number of neighborhood samples and queries.

In detail, we  transform each neighborhood sample $\bm{X}'_i$ into the binary vector $\bm{V}'_i$. If the origin word is removed in $\bm{X}'_i$, its corresponding vector dimension in $\bm{V}'_i$ is 0 otherwise 1. Therefore, $\bm{V}'_i$ has the same length as $\bm{X}'_i$, which is the length of the benign example. A benign example $\bm{X}$ is also transformed to $\bm{V}$.  Sometimes neighborhood samples may not necessarily be linearly separable,  LIME adopts gaussian kernel to weight the loss for each sample to gather points closest to the original sample, which helps with linear fitting. We  give weights $\pi(\bm{V}'_i,\bm{V})$ to each neighborhood sample according to their distance from the benign  sample~\cite{limeribeiro2016should}.
\begin{equation}
    \pi(\bm{V}'_i,\bm{V}) = \exp{(-d(\bm{V}'_i,\bm{V})^2/ \sigma^2)}
\end{equation}
where $d(\cdot,\cdot)$ is a distance function. We adopt the  cosine similarity as the distance metric.
\begin{equation}
    d(\bm{V}'_i,\bm{V}) = \frac{\bm{V}'_i\cdot \bm{V}}{\sqrt{\lvert \bm{V}'_i\rvert  \lvert \bm{V} \rvert}}
\end{equation}
Finally, we  calculate the optimal parameters $\bm{\theta^*}$:
\begin{equation}
    \bm{\theta^*} = \underset{\bm{\theta}}{\arg\min} \sum_{i=1}^n \pi(\bm{V}'_i,\bm{V}){\{\hat{\bm{Y}'_i}}-g(\bm{X}'_i)\}^2 + \Omega(\bm{\theta})
\end{equation}
where $\Omega(\bm{\theta})$ is the non-zero of parameters, which is a measure of the complexity of the linear model.  After optimizing $\bm{\theta}$, the importance of each word $x_i$ is equal to $\theta_i$. LIME can be seen as an approximation of the model's decision boundary in the original sample. The parameters can be interpreted as the margin, the larger the margin, the larger the importance of this word in approximating the decision boundary. We will filter out stop words using NLTK\footnote{https://www.nltk.org/} firstly and calculate the importance of each word.  To ensure that LimeAttack has generated high-quality adversarial examples rather than just negative examples.  We only adopt synonym replacement strategy and construct the synonym candidate set $\mathcal{C}(x_i)$ for each word $x_i$ by selecting the top $k$ nearest synonyms in the counter-fitted embedding space~\cite{mrksic-etal-2016-counter}.  Additionally,  we present the results of human evaluation and  more qualitative adversarial examples in  Appendix I.

\subsubsection{Perturbation Execution.}
\label{3.2.2}
Adversarial examples generation is a combinatorial optimization problem. 
 Score-based attack iterates by selecting the token that causes the greatest change in model's logits each time. But there is no such information in the hard-label attack. Therefore, we can only rely on the similarity between the adversarial sample and the original sample for iteration. The problem is that the similarity and attack success rate are not completely linearly correlated. As shown in the Table.\ref{tab.sample}, greedily selecting the adversarial sample with the lowest similarity each time cannot ensure that the final attack success rate is optimal. We hope that each sampling is uniformly distributed to balance attack success rate and semantic similarity.  For each origin word $x_i$, we replace it with $c \in \mathcal{C}(x_i)$ to generate an adversarial  example $\bm{X}'=[x_1,\cdots,x_{i-1},c,x_{i+1},\cdots,x_n]$,  then we calculate the  semantic similarity between  the benign sample $X$ and the adversarial example $\bm{X}'$ by universal sentence encoder (USE)\footnote{https://tfhub.dev/google/ universal-sentence-encoder}. We first sort  candidates by similarity and sample $b$ adversarial examples each time to enter the next iteration. In detail,  We have formulated the following sampling rules: (1) Sampling $\lfloor b/3 \rfloor$ adversarial examples with the highest semantic similarity. (2) Sampling $\lfloor b/3 \rfloor$ adversarial examples with the lowest semantic similarity. (3) Sampling $\lfloor b/3 \rfloor$ of the remaining adversarial samples randomly.  The analysis of hyper-parameters $b$ and  LimeAttack's algorithm are summarized in Appendix C and H.

\section{Experiments}
Analysis of the transferability and adversarial training of LimeAttack are listed in Appendix D and E.
\subsection{Tasks, Datasets and Models}

We adopt seven common datasets,  such as MR~\cite{mr_data}, SST-2~\cite{socher2013recursivesst}, AG~\cite{zhang2015character} and Yahoo~\cite{yoo-etal-2020-searching} for text classification. SNLI~\cite{bowman2015large} and MNLI~\cite{williams2018broad} for textual entailment, where MNLI includes a matched version (MNLIm) and a mismatched version (MNLImm).  In addition, we have trained three neural networks as victim models, including CNN~\cite{kim-2014-convolutional}, LSTM~\cite{6795963} and  BERT~\cite{bert2019}. The parameters of the models and the  detailed information of  datasets are listed in Appendix A.

\subsection{Baselines}
We have chosen the following existing hard-label attack algorithms as our baselines: HLBB~\cite{hlbbmaheshwary2021generating}, TextHoaxer~\cite{ye2022texthoaxer}, LeapAttack~\cite{ye2022leapattack} and TextHacker~\cite{texthackeryu2022learning} as our baselines. Additionally, we have included  some classic score-based attack algorithms, such as  TextFooler (TF)~\cite{2020-textfooler}, PWWS~\cite{2020pwds} and Bert-Attack~\cite{li-etal-2020-bert-attack} for references, which obtain additional confidence scores for attacks and are implemented on the TextAttack framework~\cite{morris2020textattack}.
\subsection{Automatic Evaluation Metrics}
We use four metrics to evaluate the attack performance: attack success rate (ASR), perturbation rate (Pert), semantic similarity (Sim) and query number (Query). Specifically, given a dataset $\mathcal{D} = \{(\bm{X}_i, \bm{Y}_i)\}_{i=1}^N$ consisting of $N$ samples $\bm{X}_i$ and corresponding ground truth labels $\bm{Y}_i$, attack success rate of an adversarial attack method $\mathcal{A}$, which generates adversarial examples $\mathcal{A}(\bm{X})$ given an input $\bm{X}$ to attack a victim model $\mathcal{F}$, is defined as~\cite{wang2021adversarial}:
\begin{equation}
ASR = \sum_{(\bm{X},\bm{Y})\in \mathcal{D}}\frac{\mathbb{I} [\mathcal{F}(\mathcal{A}(\bm{X})) \neq \bm{Y}]}{|\mathcal{D}|}
\end{equation}
The perturbation rate  is the proportion of the number of substitutions to the number of original tokens, which has been defined in Eq~\ref{eq.pert}. The semantic similarity  is measured by the  Universal Sentence Encoder (USE). Most papers~\cite{hlbbmaheshwary2021generating,ye2022leapattack}  have adopted USE. In order to maintain consistency and facilitate comparability, we have also utilized USE. Query number  is the number of  model queries  during the  attack. The robustness of a model is inversely proportional to the attack success rate, while the perturbation rate and semantic similarity together reveal the quality of adversarial examples. Query number reveals the attack efficiency.
\subsection{Implementation Details}
We set the kernel width $\sigma=25$, the number of neighborhood samples equal to the number of  the benign sample's tokens, and the beam size $b=10$. For a fair comparison, all baselines follow the same settings:  synonyms are selected from counter-fitted embedding space and  the  number of each candidate set $k=50$, the same 1000 texts are sampled for baselines to attack.   The results are averaged on five runs with different seeds (1234,2234,3234,4234 and 5234) to eliminate randomness. In order to improve the quality of adversarial examples, the attack succeeds if  the perturbation rate of each adversarial example is  less than 10\%.  We set  a tiny query budget of  100 for hard-label attack, which corresponds to real-world settings. (\eg, The HuggingFace free Inference API typically limits calls to 200 times per minute.)

\subsection{Experiments Results}

\paragraph{Attack Performance.}
Table~\ref{tab.asr} and ~\ref{tb:entail_performance} show that LimeAttack outperforms existing hard-label attacks on text classification and textual entailment tasks, achieving higher attack success rates and lower perturbation rates in datasets such as SST-2, AG, and MNLI.   Unlike existing hard-label attacks that require many queries to optimize the perturbation, LimeAttack adopts a local explainable method to calculate word importance ranking and attacks key words first. This approach can generate adversarial examples with a high attack success rate, even under tiny query budgets.  Appendix G includes a t-test and the mean and variance of LimeAttack's success rate compared to other methods.In Appendix K and L, we list the semantic similarity and the results of the comparison results between LimeAttack and several score-based attacks.

\begin{table*}[htb]
\centering
\normalsize
\begin{tabular}{clccccccccccc} 
\toprule
\multirow{2.2}*{\textbf{ Model}} & \multirow{2.4}*{\textbf{Attack}} & \multicolumn{2}{c}{\textbf{MR}} &  &
\multicolumn{2}{c}{\textbf{SST-2}} & &
\multicolumn{2}{c}{\textbf{AG}} &  &
\multicolumn{2}{c}{\textbf{Yahoo}} \\

\cmidrule(lr){3-4}\cmidrule(lr){6-7}\cmidrule(lr){9-10}\cmidrule(lr){12-13}

~ & ~   & \textbf{ASR.$\uparrow$} & \textbf{Pert.$\downarrow$}     && \textbf{ASR.$\uparrow$} & \textbf{Pert.$\downarrow$}      && \textbf{ASR.$\uparrow$} & \textbf{Pert.$\downarrow$}   && \textbf{ASR.$\uparrow$} & \textbf{Pert.$\downarrow$} \\ 

\midrule
\multirow{4}*{\textbf{CNN}}& HLBB & 44.4 & 5.4 &&  33.4 & ~~\textbf{5.6} && 17.7  & 3.3   &&41.8&~~3.6\\ 
~  & TextHoaxer &  44.2  & \textbf{5.2} && 38.1  & ~~\textbf{5.6}    &&  15.7  & \textbf{2.9}    && 39.9  & ~~\textbf{3.3}   \\

~  & LeapAttack  & 43.1  & 5.3 && 40.0  & ~~5.7   && 20.2  & 3.2   && 40.4   & ~~3.4     \\
~  & TextHacker  & 49.4  & 6.2 && 38.1  & ~~6.3   && 20.5  & 6.2   && 38.1   & ~~5.9    \\
~ & \textbf{LimeAttack}   & \textbf{49.9}  & 5.3  && \textbf{42.8} & ~~\textbf{5.6} && \textbf{20.9}  & \textbf{2.9}   && \textbf{43.7} & ~~3.7   \\

\midrule
\multirow{4}*{\shortstack[c]{\textbf{LSTM}}}& HLBB & 41.2  & \textbf{5.2}    && 33.1 & ~~5.7   & & 15.2  & 3.1     && 38.4  &  ~~\textbf{3.3}   \\
~  & TextHoaxer  & 39.3 & 5.4   && 36.4& ~~5.6    && 14.7 & \textbf{2.7}   && 37.1 & ~~\textbf{3.3}    \\

~  & LeapAttack  & 40.0 & 5.3  && 39.8 & ~~5.6    && 15.9& 3.1    && 37.6 & ~~\textbf{3.3}     \\
~  & TextHacker  & 45.8  & 6.1 && 35.2  & ~~6.4   && 16.5  & 6.2   && 36.8   & ~~5.9    \\
~ & \textbf{LimeAttack} & \textbf{47.6} &  5.4 && \textbf{40.1} &  ~~\textbf{5.5}  && \textbf{17.3} & \textbf{2.7}   && \textbf{40.3} & ~~3.7 \\

\midrule
\multirow{4}*{\shortstack[c]{\textbf{BERT}}}& HLBB & 26.6  &  5.6 & &  23.0  & ~~5.8   && 12.7  & 3.2  && 36.3  & ~~3.6  \\
~  & TextHoaxer & 27.0 & 5.5 && 24.9  & ~~5.8 &&  9.8 & 3.0 && 32.7  & ~~\textbf{3.3} \\

~  & LeapAttack  & 26.5 & \textbf{5.4} &&  26.1 & ~~5.8 && 13.7 & \textbf{2.9} &&  34.1 & ~~3.4  \\
~  & TextHacker  & 26.5  & 6.5 && 25.4  & ~~6.3   && 12.9  & 5.5   && 31.3   & ~~6.3  \\
~ & \textbf{LimeAttack} & \textbf{29.2}  & 5.9 && \textbf{27.8}  & ~~\textbf{5.7} && \textbf{14.6}  & \textbf{2.9} && \textbf{37.4}  & ~~3.8\\
\bottomrule
\end{tabular}

\caption{The attack success rate (ASR.,\%$\uparrow$) and perturbation rate (Pert.,\%$\downarrow$) of different hard-label attack algorithms on three models for text classification under a query budget of 100.}
\label{tab.asr}
\end{table*}

\begin{table*}[htb]
\normalsize
\centering
\begin{tabular}{ccccccccccc}
\toprule
\multirow{2}{*}{\textbf{Dataset}} & \multicolumn{2}{c}{\textbf{HLBB}} & \multicolumn{2}{c}{\textbf{TextHoaxer}} & \multicolumn{2}{c}{\textbf{LeapAttack}} & \multicolumn{2}{c}{\textbf{TextHacker}} & \multicolumn{2}{c}{\textbf{LimeAttack}} \\\cmidrule(lr){2-3}\cmidrule(lr){4-5}\cmidrule(lr){6-7}\cmidrule(lr){8-9}\cmidrule(lr){10-11}
& \textbf{ASR.$\uparrow$}    &\textbf{Pert.$\downarrow$}           & \textbf{ASR.$\uparrow$}    &\textbf{Pert.$\downarrow$}             & \textbf{ASR.$\uparrow$}    &\textbf{Pert.$\downarrow$}             & \textbf{ASR.$\uparrow$}    &\textbf{Pert.$\downarrow$}             & \textbf{ASR.$\uparrow$}    &\textbf{Pert.$\downarrow$}         \\\midrule
\textbf{SNLI}                     & 24.9    & \textbf{8.3}   & 24.7       & \textbf{8.3}      & 28.3       & \textbf{8.3}      & 22.8       & \textbf{8.3}      & \textbf{29.1}  & 8.4           \\
\textbf{MNLIm}                     & 41.9    & 7.8            & 40.9       & \textbf{7.7}      & 49.1       & \textbf{7.7}      & 38.2       & 7.8               & \textbf{49.7}  & \textbf{7.7}  \\
\textbf{MNLImm}                   & 47.8    & \textbf{7.5}   & 45.6       & 7.6               & 56.0         & 7.6               & 44.3       & 7.7               & \textbf{56.3}  & 7.6          \\\bottomrule
\end{tabular}
\caption{The ASR.,\%$\uparrow$ and Pert.,\%$\downarrow$ of LimeAttack and other baselines on BERT for  textual entailment under a query budget of 100.}
\label{tb:entail_performance}
\end{table*}

\paragraph{Query Budget.}
As illustrated in Figure~\ref{fig.diff}, LimeAttack still maintains a stable attack success rate and a smoother attack curve under different query budgets, which means that regardless of high or low query budget, LimeAttack often have a stable and excellent attack performance. The trend of perturbation rate are listed in Appendix N.  Comparing the attack performance in low query and high query budgets can provide a more comprehensive evaluation. However, attack without considering the query budget is more of an ideal situation, it shows the upper limit of an attack algorithm. A large number of queries are expensive, we believe attack performance under low query budget is more practical.   We also list some attack success rates and perturbation rates of different attacks under the query budget is 2000 in Appendix N.

\begin{figure}[htb]
\centering
\includegraphics[width=\linewidth]
{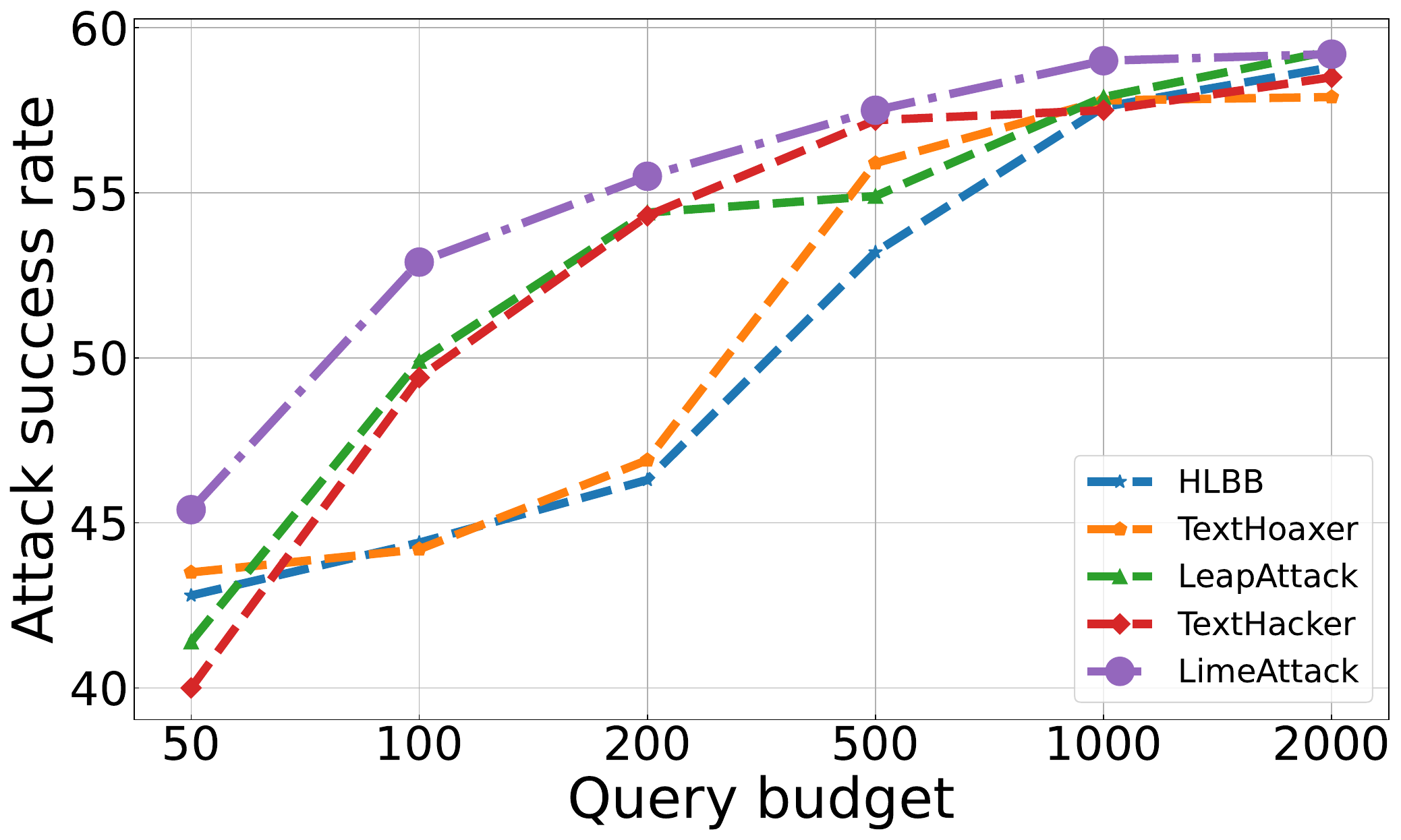} 
\caption{ Attack success rate of different attacks under different query budgets  on CNN-MR.}
\label{fig.diff}
\end{figure}

\paragraph{Adversary Quality.} 
High-quality adversarial examples should be both fluent and context-aware, while also being similar to benign samples to evade human detection. We utilize Language-Tools\footnote{https://www.languagetool.org/} and USE to detect grammatical errors and measure semantic similarity. As shown in Table~\ref{tab.semantic}, LimeAttack has the lowest perturbation rate and grammatical error, though its semantic similarity is lower than HLBB, TextHoaxer, and LeapAttack. Because these methods take the similarity into account during the attack, thus LimeAttack exhibits lower similarity than other methods. Considering all metrics, LimeAttack is still dominant.  To intuitively contrast the quality of adversarial examples, some qualitative examples are provided in Appendix I.

\begin{table}[h]
\normalsize
\begin{center}
\begin{tabular}{lcccc}
\toprule

{\textbf{Attack}} & \textbf{ASR.$\uparrow$} & \textbf{Pert.$\downarrow$} & \textbf{Sim.$\uparrow$}  & \textbf{Gram.$\downarrow$}\\

\midrule
HLBB  & 23.0 & 5.8 & 99.2 &  1.6 \\
TextHoaxer  & 24.9 & 5.8 & \textbf{99.2} & 1.7 \\
LeapAttack & 26.1 & 5.8 & 99.1 & 1.5 \\
TextHacker  & 25.4 & 6.3 & 96.0 & 1.9 \\
\textbf{LimeAttack} & \textbf{27.8} & \textbf{5.7} & 96.4 & \textbf{1.5}  \\
\bottomrule
\end{tabular}
\end{center}
\caption{ASR.,\%$\uparrow$, Pert.,\%$\downarrow$, Sim.,\%$\uparrow$ and Gram.,$\downarrow$ of different hard-label attack algorithms on SST-2 dataset for  BERT under query budget of 100.}
\label{tab.semantic}
\end{table}

\paragraph{Evaluation on Large Language Models.} 
\begin{table}[h]
\begin{center}
\normalsize
\begin{tabular}{lllll}
\toprule
\textbf{Model(size)}           & \textbf{ASR.$\uparrow$}  & \textbf{Pert.$\downarrow$} & \textbf{Sim.$\uparrow$}  & \textbf{Acc.$\uparrow$} \\
\midrule
BART-L (407M) & 42.0   & 5.15 & 93.7 & 87.0     \\
DeBERTa-L (435M) &52.0   &5.82      & 92.9     &79.0          \\
T5-L (780M)   &28.0&5.59&95.1&\textbf{93.0}\\
GPT3(175B) & \textbf{61.0} & \textbf{4.82} & 95.2 & 82.0     \\
ChatGPT (175B)  &25.0 &    5.62  &  \textbf{95.3}   & 92.0    \\\bottomrule
\end{tabular}
\end{center}
\caption{The evaluation of LimeAttack on large language models. We attack these large language models on MR dataset under query budget of 100.}
\label{tab.llm}
\end{table}    

Large language models (LLMs), also known as foundation models~\cite{bommasani2021opportunities}, have achieved impressive performance on various natural language processing tasks. However, their robustness to adversarial examples remains unclear~\cite{wang2023robustness}. To evaluate the effectiveness of LimeAttack on LLMs, we select some popular models such as DeBERTa-L~\cite{deberta}, BART-L~\cite{lewis2019bart}, Flan-T5~\cite{T5}, GPT-3 (text-davinci-003) and ChatGPT (gpt-3.5-turbo)~\cite{gpt3}. Due to the limited API calls, we sample 100 texts from MR datasets and attacked the zero-shot classification task of these models. As Table~\ref{tab.llm} shows, LimeAttack successfully attacked most LLMs under tight query budgets. Although these models have high accuracy on zero-shot tasks, their robustness to adversarial examples still needs to be improved. ChatGPT and T5-L are more robust to adversarial examples. The robustness of the victim model is related to origin accuracy. The higher the origin accuracy, the stronger the victim model's ability to defense adversarial examples.  Further analysis of other hard-label attacks and experimental details are discussed in Appendix F.

\begin{table*}[htb]
\normalsize
\centering
\begin{tabular}{ccccccccccc}
\toprule
\multirow{2}{*}{\textbf{Defense Method}} & \multicolumn{2}{c}{\textbf{HLBB}} & \multicolumn{2}{c}{\textbf{TextHoaxer}} & \multicolumn{2}{c}{\textbf{LeapAttack}} & \multicolumn{2}{c}{\textbf{TextHacker}} & \multicolumn{2}{c}{\textbf{LimeAttack}} \\\cmidrule(lr){2-3}\cmidrule(lr){4-5}\cmidrule(lr){6-7}\cmidrule(lr){8-9}\cmidrule(lr){10-11}
& \textbf{ASR.$\uparrow$}    &\textbf{Pert.$\downarrow$}           & \textbf{ASR.$\uparrow$}    &\textbf{Pert.$\downarrow$}             & \textbf{ASR.$\uparrow$}    &\textbf{Pert.$\downarrow$}             & \textbf{ASR.$\uparrow$}    &\textbf{Pert.$\downarrow$}             & \textbf{ASR.$\uparrow$}    &\textbf{Pert.$\downarrow$}         \\\midrule
None            & 24.9    & \textbf{8.3}   & 24.7       & \textbf{8.3}      & 28.3       & \textbf{8.3}      & 22.8       & \textbf{8.3}      & \textbf{29.1}  & 8.4           \\
A2T               & 20.6    & 9.3            & 21.4       & \textbf{9.5}      & 23.5       & \textbf{9.4}      & 19.8       & \textbf{9.1}      & \textbf{24.5}  & \textbf{9.4}  \\
ASCC              & 13.2    & \textbf{6.5}   & 13.4       & 6.5               & 14.3       & \textbf{6.4}      & 12.5       & 7.2               & \textbf{15.8}  & 6.7      \\\bottomrule    
\end{tabular}
\caption{The evaluation of hard-label attacks on defense methods based on BERT-SNLI under query budget of 100.}
\label{tab.dense}
\end{table*}
 \paragraph{Attack Performance on Defense Methods.}
 To evaluate the effectiveness of LimeAttack on defense methods, we use A2T~\cite{yoo} and ASCC~\cite{dong2021towards} to enhance the defense ability of BERT on SNLI, and conducted attack experiments on this basis. As shown in Table~\ref{tab.dense}, LimeAttack still has a certain attack effect and outcomes other baselines on these defense methods. More attack performance on defense methods are listed in Appendix M.

\subsection{Ablation Study}
\paragraph{Effect of Word Importance Ranking.}
To validate the effectiveness of word importance ranking, we removed the word importance ranking strategy and instead randomly selected words to perturb to evaluate its effectiveness. Table~\ref{tab.wir} shows that without the word importance ranking, the attack success rate decreased by 9\% and 6\% on the MR and SST-2 datasets, respectively. Furthermore, adversarial examples generated by random selection had higher perturbation rates and required more queries. This indicates the importance of the word importance ranking in guiding LimeAttack to focus on crucial words, leading to a more  efficient attack with lower perturbation rates.

\paragraph{Effect of Sampling Rules.}
To verify the effectiveness of LimeAttack's sampling rules, we will replace this strategy with one of three common sampling rules: (1) selecting $b$ adversarial examples with the highest semantic similarity, (2) selecting $b$ adversarial examples with the lowest semantic similarity, or (3) randomly selecting $b$ adversarial examples. The results in Table~\ref{tab.sample} show that LimeAttack outperforms other sampling rules with a higher attack success rate and lower perturbation rate.  Additionally, it has a comparable (second highest) semantic similarity and number of queries.

\begin{table}[H]
\normalsize
\centering
\begin{tabular}{ccccc}
\toprule
\multirow{2}{*}{} & \multicolumn{2}{c}{\textbf{MR}} & \multicolumn{2}{c}{\textbf{SST-2}} \\\cmidrule(lr){2-3}\cmidrule(lr){4-5}
                  & \textbf{Random}     & \textbf{LIME}      & \textbf{Random}       & \textbf{LIME}       \\
                  \midrule
\textbf{Pert.$\downarrow$}              & 6.1        & \textbf{5.6}       & 6.4          & \textbf{5.9}        \\
\textbf{ASR.$\uparrow$}               & 30.1         & \textbf{39.3}      & 32.1         & \textbf{36.5}       \\
\textbf{Sim.$\uparrow$}              & 94.6       & \textbf{94.8}      & 94.2         & \textbf{94.6}       \\
\textbf{Query.$\downarrow$}             & 157.2      & \textbf{153.3}     & 148.1        & \textbf{132.5}      \\\bottomrule
\end{tabular}
\caption{Comparison between word importance ranking learned by LIME and random selecting  for BERT under query budget of 1000.}
\label{tab.wir}
\end{table}

\begin{table}[H]
\normalsize
\centering
\begin{tabular}{ccccccc}
\toprule
\textbf{Sample Rule}       & \textbf{ASR.$\uparrow$} & \textbf{Pert.$\downarrow$} & \textbf{Sim.$\uparrow$}  & \textbf{Query.$\downarrow$} \\\midrule
Method 1         & 35.8            & 5.76           & \textbf{95.02} & 164.65          \\
Method 2         & 31.5            & 6.13           & 93.79          & \textbf{87.45}  \\
Method 3        & 32.1           & 6.09           & 94.50           & 107.05          \\
\textbf{LimeAttack} & \textbf{39.3}   & \textbf{5.65}  & 94.81          & 153.03          \\\bottomrule
\end{tabular}
\caption{Comparison between different sample rules on MR dataset for BERT under query budget of 1000.}
\label{tab.sample}
\end{table}

\subsection{Human Evaluation}
\label{sec.eval}
We selected 200 adversarial examples BERT-MR. Each adversarial example was evaluated by two human judges for semantic similarity, fluency and prediction accuracy. The entire human evaluation is consistent with TextFooler~\cite{2020-textfooler}.   In detail, we ask human judges to put a  5-point Likart scale (1-5 corresponds to very not fluent/similar, not fluent/similar, uncertain, fluent/similar, very fluent/similar respectively) to evaluate the the similarity and fluency of adversarial examples and benign samples.  The results are listed in the Table~\ref{tab.human}, semantic similarity is 4.5, which  means adversarial samples are similar to original sample.  The prediction accuracy here is to make humans to predict  what the label of this sentence is (such as it is positive or negative for sentiment analysis). 76.7\% means majorities of adversarial examples have the same attribute as original samples from humans’ perspective but mistake victim model. 
\begin{table}[H]
\normalsize
\begin{center}
\begin{tabular}{llll}
\toprule
            & \textbf{Ori} && \textbf{Adv} \\\midrule
\textbf{Prediction Accuracy} & 81.2\%              && 76.7\%                    \\
\textbf{Fluency}             & 4.4              && 4.1                    \\
\textbf{Semantic Similarity}         &&4.5&                               \\\bottomrule
\end{tabular}
\end{center}
\caption{The semantic similarity, fluency and prediction accuracy  of original texts and adversarial examples evaluated by human judges for BERT-MR.} 
\label{tab.human}

\end{table}

\section{Conclusion}
In this work, we summarize the previous score-based attacks and hard-label attacks and propose a novel hard-label attack algorithm called LimeAttack. LimeAttack adopts a local explainable method to approximate the word importance ranking, and then utilizes beam search to generate high-quality adversarial examples with tiny query budget. Experiments show that LimeAttack achieves a higher attack success rate than other hard-label attacks. In addition, we have evaluated LimeAttack's attack performance on large language models and some defense methods. The adversarial examples crafted by LimeAttack are high-quality, high transferable and improves victim model's robustness in adversarial training.   LimeAttack has verified the effectiveness of inside-to-outside attack path in hard-label. Then many excellent score-based attacks may provide hard-label attacks more insight.

\bibliography{ref}
\newpage
\appendix
\section{Appendix A: Victim Model and Datasets}
In our experiments, we carry out all experiments on  NVIDIA Tesla V100 16G GPU. 
We adopt three neural networks CNN,LSTM and BERT from TextFooler. The CNN consists of three window sizes of 3, 4, and 5, and 100 filters for each window size. The LSTM consists of a bidirectional LSTM layer with 150 hidden states. Both CNN and LSTM have a dropout rate of 0.3 and 200-dimensional Glove word embeddings pre-trained on 6B tokens. The BERT$_{base}$ consists of 12 layers with 768 units  and 12 heads. The origin accuracy of victim models are listed in Table~\ref{tab.acc}. Detailed datasets are listed in Table~\ref{tab.dataset}. We select different text length and different classes datasets.

        \begin{table}[h]
            \centering
            \begin{minipage}[t]{0.48\linewidth}
                \caption{The original accuracy of victim model on various data sets.}
                \label{tab.acc}
                \resizebox{\textwidth}{!}{
                    \begin{tabular}{lcccccc}
                        \toprule
                    \textbf{Dataset} & \textbf{} & \textbf{CNN} & \textbf{} & \textbf{LSTM} & \textbf{} & \textbf{BERT} \\\midrule
                    MR               &           & 78.0           &           & 80.7          &           & 86.0            \\
                    SST-2            &           & 82.7         &           & 84.5          &           & 92.4          \\
                    AG               &           & 91.5         &           & 91.3          &           & 94.2          \\
                    Yahoo            &           & 73.7         &           & 73.7          &           & 79.1          \\\midrule
                    SNLI             &           & -             &           &-               &           & 89.1          \\
                    MNLIm            &           &-              &           &-               &           & 85.1          \\
                    MNLImm           &           &-              &           &-               &           & 82.1     \\\bottomrule    
                    \end{tabular}}
            \end{minipage}
        \hfill
            \begin{minipage}[t]{0.48\linewidth}
                \caption{Overview of datasets and NLP tasks.}
                \label{tab.dataset}
                \resizebox{\textwidth}{!}{
                    \begin{tabular}{c|ccccc}\midrule
                        \textbf{Task}                            & \textbf{Dataset}   & \textbf{Train} & \textbf{Test} &\textbf{Classes} & \textbf{Length} \\ \midrule
                        \multirow{4}{*}{Classification} & MR        & 9K    & 1K  &2 & 18  \\
                                                        & SST-2     & 70K   & 2K  &2  & 8  \\
                                                        &AG     &120K  &8K   &4 &43  \\ 
                                                        &Yahoo      &12K &4K    &10   &151 \\\midrule
                        \multirow{2}{*}{Entailment}     & SNLI      & 570K  & 3K &3  & 20  \\
                                                        & MNLI(m/mm)  & 433K  & 10K &3  & 11  \\\bottomrule
                        \end{tabular}}
            \end{minipage}
        \end{table}

\section{Appendix B: The Effectiveness of LIME in Score-based Attacks}
Traditional score-based attacks utilize deletion-based methods to calculate word importance ranking. They drop a word $x_i$ from the benign sample $X$ and query the victim model $\mathcal{F}$ with the new sample $X/x_i=[x_1,x_2,\cdots,x_{i-1},x_{i+1},\cdots,x_n]$. The difference in the model's confidence score before and after deletion reflects the importance of this word:
\begin{equation}
    I(x_i) = \mathcal{F}(X) - \mathcal{F}(X/x_i)
\end{equation}
To verify the effectiveness of local explainable method, we replace deletion-based method with local explainable method in the score-based attack. We test on MR data set and  results are shown in the Table~\ref{tab.score}.   Local explainable method  and deletion-based method achieve similar attack success rate, but deletion-based method achieves lower perturbation rate than local explainable method.  Because  the probability distribution of the model's output is available, the influence of each word on the output can be well reflected by deletion-based method. Therefore, compared with score-based attacks, we think local  explainable methods can play a greater advantage in hard-label attacks where deletion-based method is useless.

\begin{table}[htb]
\Large
    \centering
    \caption{The comparison with deletion-based method. ASR.,\%$\uparrow$ is attack success rate and  Pert.,\%$\downarrow$ is perturbation rate.}
    \label{tab.score}
    \resizebox{\linewidth}{!}{
    \begin{tabular}{ccccccc}
        \toprule
    \multirow{2}{*}{\textbf{Dataset}} & \multirow{2}{*}{\textbf{\makecell[c]{Victim\\Models}}} & \multicolumn{2}{c}{\textbf{Deletion-based}} & \multicolumn{2}{c}{\textbf{LIME}} \\\cmidrule{3-6}
                                      &                                        & \textbf{ASR.$\uparrow$}       & \textbf{Pert.$\downarrow$}       & \textbf{ASR.$\uparrow$}             & \textbf{Pert.$\downarrow$}             \\\midrule
    \multirow{3}{*}{MR}               & CNN                                    & \textbf{1.0}                    & \textbf{11.9}                 & \textbf{1.0}                          & 12.4                       \\
                                      & LSTM                                   & \textbf{0.6}                  &\textbf{12.3}                & \textbf{0.6}                        & 12.8                       \\
                                      & BERT                                   & 8.2                  & \textbf{16.3}                 & \textbf{8.1}                        & 17.4                       \\\bottomrule               
    \end{tabular}}
    \end{table}

%To verify the effectiveness of such learned word importance ranking, we  compare  with  word importance ranking calculated by score-based attack.  We adopt cosine similarity to  measure the similarity between both. As shown in the Tab.~\ref{tab.cos}, the word importance ranking calculated by local explainable method can maintain more than 75\% similarity with the score-based attack on MR, SST-2 and Yahoo data sets.  This means we can effectively use local explainable method to approximate the importance ranking of words in a hard-label setting.

%\begin{table}[h]
%\centering
%\begin{tabular}{ccccc}
%\toprule
%\textbf{Dataset} & \textbf{Avg. Length} & \textbf{CNN} & \textbf{LSTM} & \textbf{BERT} \\\midrule
%MR               & 20                  & 80.5         & 81.1          & 76.4          \\
%SST-2            & 8                   & 78.9         & 78.2          & 76.4          \\
%Yahoo            & 150                 & 76.4         & 78.5          & 77.1         \\\bottomrule
%\end{tabular}
%\caption{The similarity between word importance ranking  calculated by  LIME and  score-based attack respectively. Avg.Length is average length of data set.}
%\label{tab.cos}
%\end{table}
\section{Appendix C: The Effectiveness of Beam Size $b$}
\begin{figure*}[h]
        \centering
        \subfigure[Attack success rate.]
        {
            \begin{minipage}[b]{.3\linewidth}
                \centering
                \includegraphics[width=\linewidth]{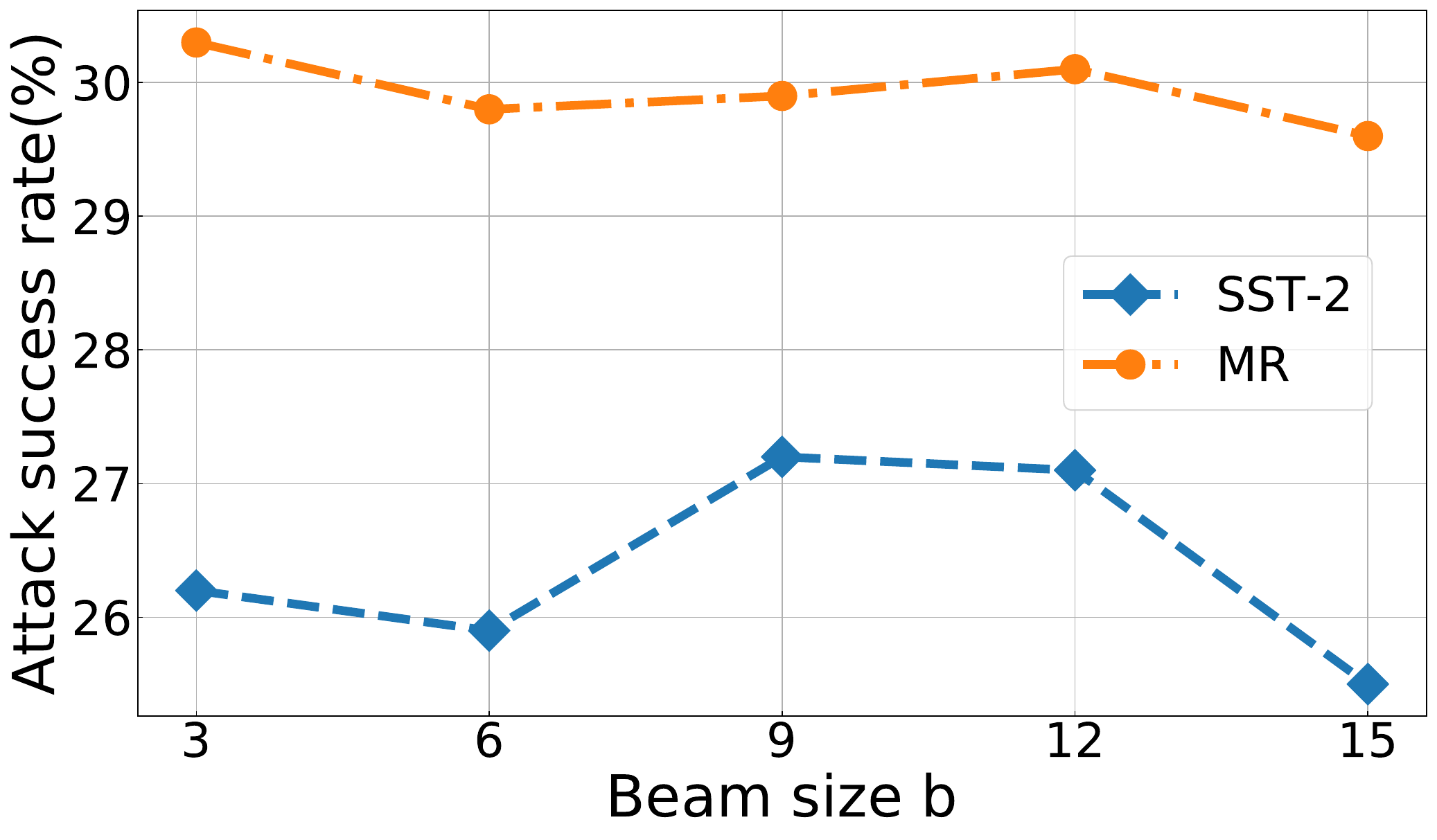} 
            \end{minipage}
            \label{figure.cnn_query}
        }
        \subfigure[Perturbation rate.]
        {
            \begin{minipage}[b]{.3\linewidth}
                \centering
                \includegraphics[width=\linewidth]{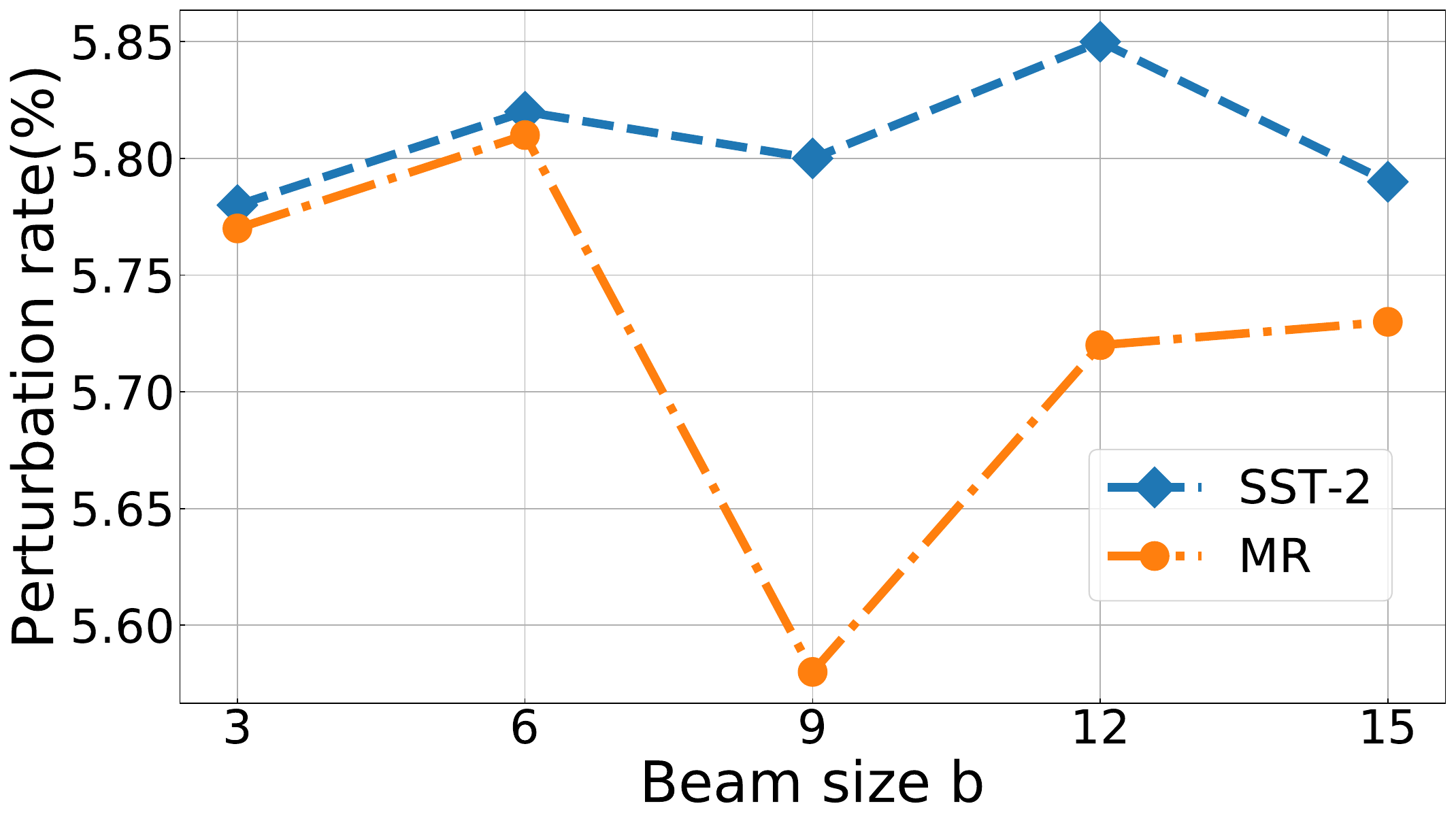} 
            \end{minipage}
            \label{figure.lstm_query}
        }
        \subfigure[Semantic similarity.]
        {
            \begin{minipage}[b]{.3\linewidth}
                \centering
                \includegraphics[width=\linewidth]{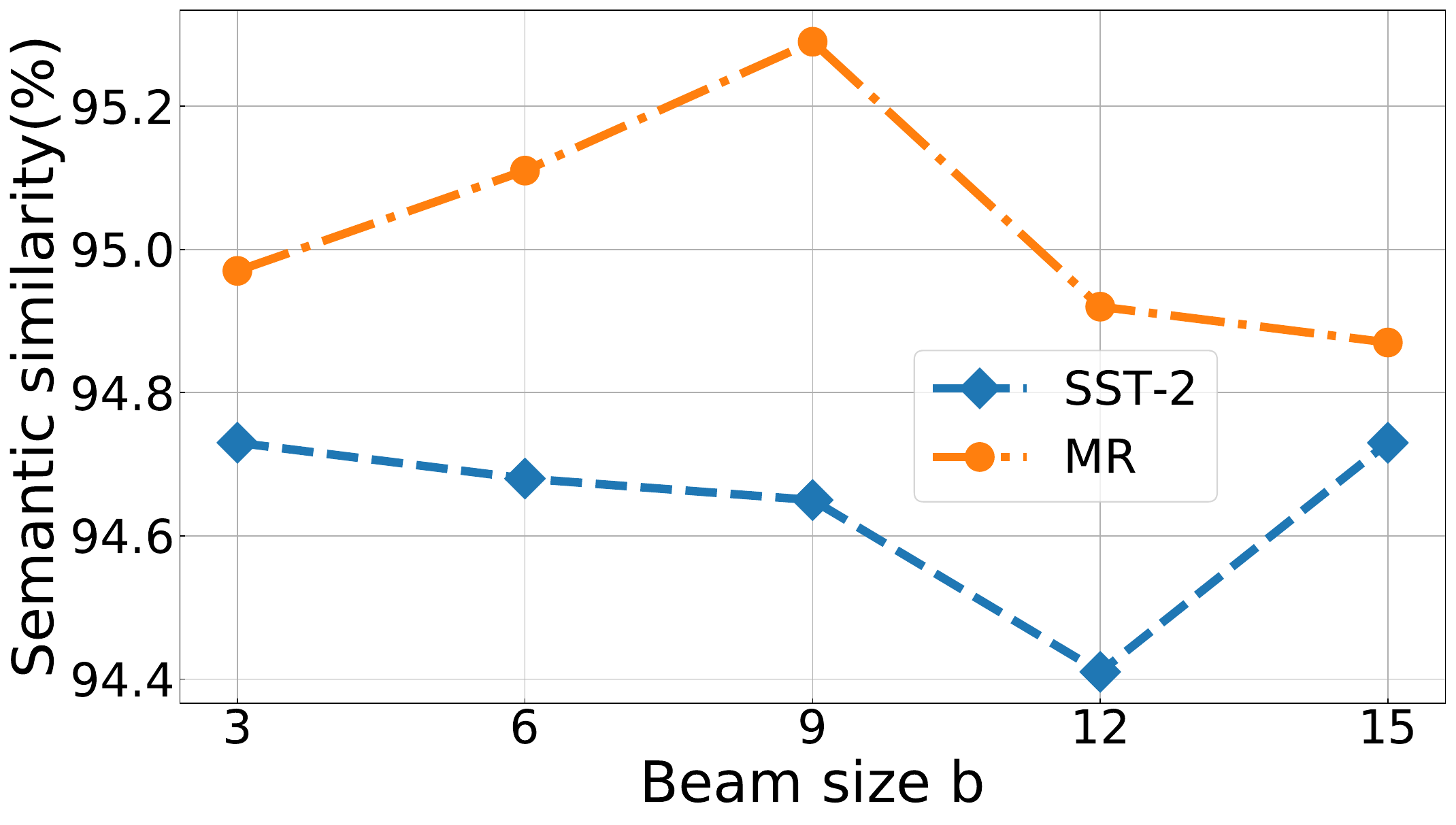}
            \end{minipage}
            \label{figure.bert_query}
        }
        \caption{The attack success rate (\%) $\uparrow$, perturbation rate (\%) $\downarrow$ and semantic similarity(\%) $\uparrow$ LimeAttack  on BERT using MR and SST-2 dataset under different beam size $b$}
        \label{fig.beamsize}
        \end{figure*}

Beam size $b$ directly determines the size of search space. Bigger search space is significant to generate the optimal solution (\eg, lower perturbation rate and higher semantic similarity), while it also requires a lot of model queries.  Therefore, how to select  an appropriate beam size to balance the query and attack success rate. As shown in the Figure ~\ref{fig.beamsize}, We test on MR and SST-2 data sets using BERT with different beam size. 
With the increase of beam size $b$, the search space is effectively expanded, and the attack success rate and the quality of adversarial examples (the perturbation rate is reduced) are improved. With the further increase of beam size $b$, the query also gradually increases, resulting in the decrease of attack success rate. Considering the comprehensive effect, we set the beam size $b=10$.

\section{Appendix D: Transferability}
\begin{figure}[h]
    \centering
    \includegraphics[scale=0.45]{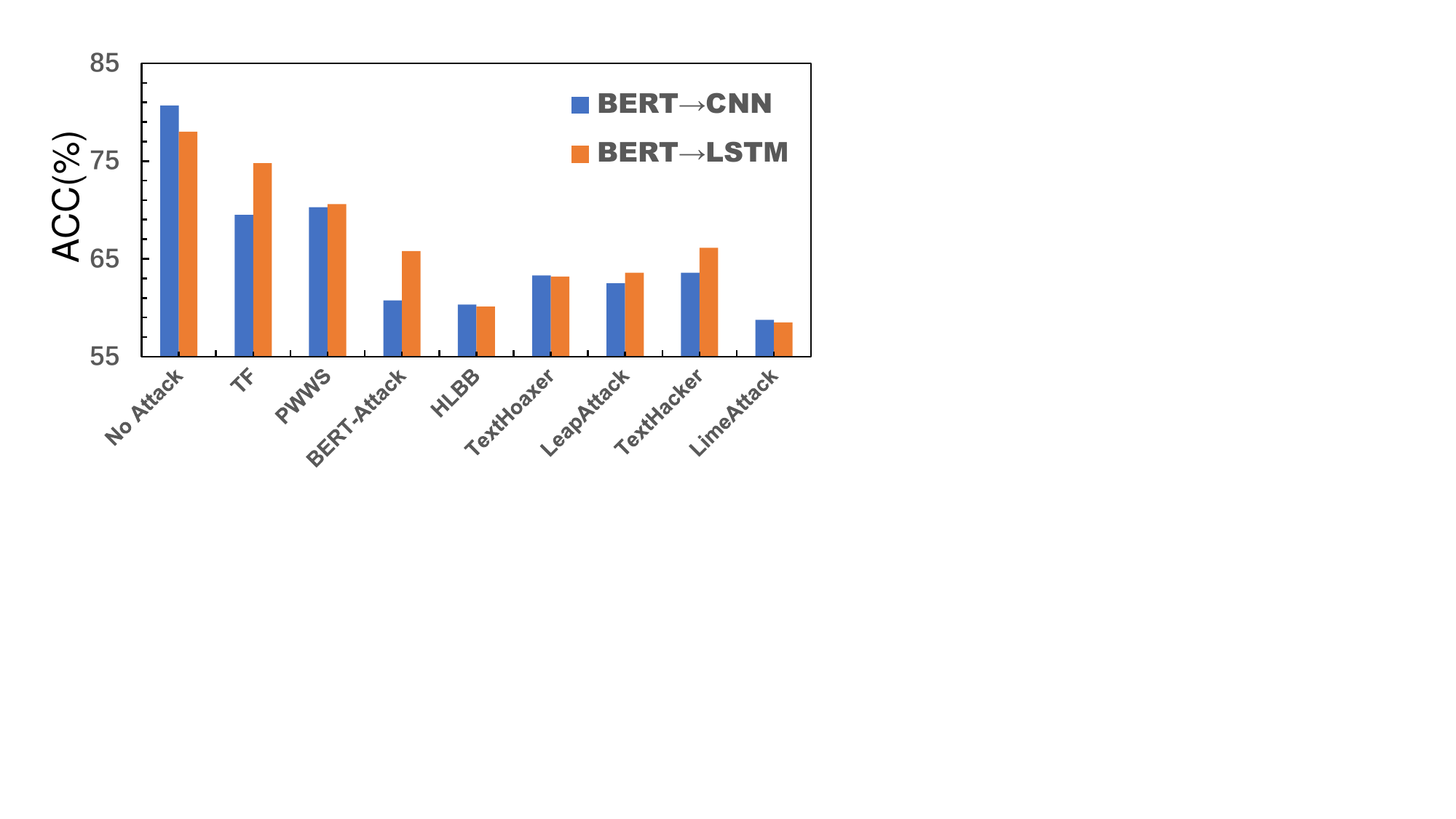}%mr_bert.pdf}
    \caption{Transferability of  adversarial examples on MR dataset for BERT. Lower accuracy indicates higher transferability.}
    \label{fig.trans}
    \end{figure}

The transferability of adversarial examples reveals the property that adversarial examples crafted by a particular victim model can also fool another. In detail, we calculate the prediction accuracy against the CNN and LSTM models on adversarial examples crafted for attacking BERT on MR dataset. As shown in the Figure~\ref{fig.trans}, adversarial examples generated by LimeAttack achieves higher transferability than baselines. It reduces the prediction accuracy of CNN and LSTM models from 80.7\%,78.0\% to 58.5\%, 58.4\% respectively.

\section{Appendix E: Adversarial Training}
Adversarial training is a prevalent technique to improve the victim model's robustness by adding adversarial examples into the training data. We randomly selected 1000 adversarial examples from the MR dataset, retrained the CNN model, and then attacked the CNN model again. The results are shown in the Table~\ref{tab.adv}, after adversarial training, the CNN model achieves higher test accuracy. In addition, LimeAttack's attack success rate has decreased by 3\% with the cost of more queries and a higher perturbation rate. Adversarial examples generated by LimeAttack effectively improve the victim model's robustness and generalization.
\begin{table}[htb]
    \centering
    \caption{The performance of CNN model with(out) adversarial training on the MR dataset. } 
    \resizebox{\linewidth}{!}{
    \begin{tabular}{cccccc}
    \toprule
    \textbf{}              & \textbf{Ori Acc.$\uparrow$} & \textbf{ASR.$\uparrow$} & \textbf{Pert.$\downarrow$} & \textbf{Sim.$\uparrow$}  & \textbf{Query.$\downarrow$} \\\midrule
    Original               & 80.27           & 38.18          & 3.90            & 97.00             & 22.21           \\
    \textbf{+Adv.Training} & 81.53  & 35.09 & 3.94  & 97.01 & 24.90  \\\bottomrule
    \end{tabular}}
    \label{tab.adv}
    \end{table}

\section{Appendix F: Large Language Models}
\subsection{Settings}
In this section, we provide a brief introduction to the large language models used in our experiments.
\begin{itemize}
\item \textbf{BART-L} BART is a transformer-based model that can handle both generation and understanding tasks. It is trained on a combination of auto-regressive and denoising objectives,  which is primarily focused on understanding tasks.
\item \textbf{DeBERTa-L} DeBERTa enhances BERT with a disentangled attention mechanism and an improved decoding scheme. This allows it to capture contextual information between different tokens more effectively and generate higher quality natural language sentences.
\item \textbf{Flan-T5} Flan-T5 uses a text-to-text approach where both input and output are natural language sentences, enabling it to perform a variety of tasks including text generation, summarization, and classification. By taking an input sentence as a prompt, Flan-T5 can accomplish common NLP tasks.
\item \textbf{Text-davinci-003 and ChatGPT} are based on GPT3 and GPT3.5. They can perform any task by natural language inputs and produce higher quality and more faithful output.
\end{itemize}
In order to ensure the stability of the output of large language models, we use the same prompt for each models under zero-shot text classification task: \textbf{Please classify the following sentence into either positive or negative. Answer me with "positive" or "negative", just one word.}
\subsection{Discuss}
\paragraph{Generalization Error.} In this subsection, we provide some analysis of models' generalization error. which is also known as the out-of-sample error. It is a measure of how accurately an algorithm is able to predict outcome values for previously unseen data.  Let $\mathcal{F}$ is a finite hypothesis set, $m$ is the number of training samples, for each $f \in \mathcal{F}$, probably approximately correct (PAC) theory reveals that:
\begin{equation}
    P \Bigg ( |\mathbb{E}(f)-\hat{\mathbb{E}}(f)| \leq \sqrt{\frac{ln|\mathcal{F}|+\varsigma }{2m}} \Bigg ) \geq 1-\delta 
\end{equation}
where $\mathbb{E}(f)$ and $\hat{\mathbb{E}}(f)$ are the ideal and empirical risk on  classifier $f$. According to the Table 6 in the main text, the robustness of the victim model is related to origin accuracy. The higher the origin accuracy, the stronger the victim model's ability to defense adversarial examples.  Generalization error relies on two factors: the training sample size ($m$) and the hypothesis space ($\mathcal{F}$). Large language models, like ChatGPT, excel in performance due to their extensive training data (large $m$). Moreover, although the hypothesis set ($\mathcal{F}$) is finite, increasing $m$ and $|\mathcal{F}|$ can lead to reduced generalization errors. This observation helps elucidate why such models excel in zero-shot classification for certain tasks.

\paragraph{Attack ChatGPT.}
To validate the attack effectiveness of  hard-label attack algorithms in the real world, we evaluate the attack performance of LimeAttack, HLBB, LeapAttack, TextHoaxer and TextHacker on ChatGPT.  Due to OpenAI's limit on the number of APIs calls, we select 20 adversarial examples generated by different hard-label attack algorithms which attack bert on the MR dataset, and input them into ChatGPT to observe if they produced opposite results compared to the original samples.  As shown in Table~\ref{chatgpt}, LimeAttack achieves higher attack success rate, generates higher quality adversarial examples than other methods when facing real world APIs under tight query budget.

\begin{table}[htb]
    \centering
    \small   
    \caption{Attack success rate (ASR., \%), perturbation rate (Pert., \%),  semantic similarity (Sim., \%) of various hard-label attacks on ChatGPT under the query budget of 100.}
    \label{chatgpt}
    \begin{tabular}{cccc}
        \toprule
    \textbf{Attack}     & \textbf{ASR.}$\uparrow$ & \textbf{Pert.}$\downarrow$ & \textbf{Sim.}$\uparrow$  \\\midrule
    HLBB       & 10.0   & \textbf{3.70}   & \textbf{96.80}  \\
    LeapAttack & \textbf{20.0}   & 8.57  & 88.85 \\
    TextHoaxer & 10.0   & 4.61  & 89.71 \\
    TextHacker & \textbf{20.0}   & 7.61  & 90.21 \\
    \textbf{LimeAttack} & \textbf{20.0}   & 4.51  & 95.30  \\\bottomrule
    \end{tabular}
    \end{table}

\section{Appendix G:  Significance Test}
We have added a t-test and  listed the mean, variance, and p-value of LimeAttack against other methods on the success rate in the Table~\ref{ttest}.  LimeAttack has run with five additional seeds and take the average, which is consistent with other baselines. As shown in the Table~\ref{ttest}, LimeAttack has achieved better results than other baselines under a tight query budget.
\begin{table*}[htb]
    \centering 
    \caption{The mean, variance, and p-value of LimeAttack against other methods on the success rate in 5 runs.}
    \label{ttest}
    %\resizebox{\linewidth}{!}{
    \begin{tabular}{ccccccc}
    \toprule
    \multirow{2}{*}{Model\_dataset} & \multicolumn{2}{c}{LimeAttack} & HLBB     & TextHoaxer     & LeapAttack    & TextHacker     \\
                    & Mean      & Variance     & p-value  & p-value  & p-value  & p-value  \\\midrule
    CNN\_MR           & 49.9      & 9.00E-02     & 2.74E-05 & 2.38E-05 & 1.19E-05 & 8.20E-02 \\
    LSTM\_MR         & 47.6      & 2.50E-01     & 8.98E-05 & 3.45E-05 & 4.78E-05 & 5.69E-03 \\
    BERT\_MR           & 29.2      & 1.42E-01     & 2.85E-02 & 6.70E-02 & 2.37E-02 & 2.37E-02 \\
    CNN\_SST          & 42.8      & 2.91E-01     & 1.58E-05 & 1.24E-04 & 4.42E-04 & 1.24E-04 \\
    LSTM\_SST         & 40.1      & 8.02E-01     & 1.53E-03 & 5.66E-03 & 6.41E-02 & 3.30E-03 \\
    BERT\_SST          & 27.8      & 4.22E-02     & 1.73E-05 & 3.39E-05 & 5.71E-05 & 4.17E-05 \\
    CNN\_AG           &20.9       &2.28E-01      &1.38E-02          &2.27E-03          &6.30E-01  &3.09E-01          \\
    LSTM\_AG          &17.3      &5.18E-02  &1.50E-03           &5.07E-04          &1.38E-02          &3.17E-01          \\
    BERT\_AG           &14.6           &1.02E-02              &4.41E-03          &5.16E-04          &3.77E-02          &5.86E-03          \\
    CNN\_Yahoo        &43.7 &1.56E-01              &6.30E-02          &1.26E-03          &2.68E-03          &1.82E-04          \\
    LSTM\_Yahoo       &40.3           &4.22E-02              &1.39E-03          &3.45E-04          &5.49E-04          &2.69E-04          \\
    BERT\_Yahoo        &37.4           &2.25E-02              &4.22E-01          &1.59E-03          &4.04E-03          &8.47E-04         \\\bottomrule
    \end{tabular}
    \end{table*}

    \section{Appendix H:  LimeAttack Algorithm}

    The all process of LimeAttack's algorithm is summarized in algo~\ref{alg:algorithm}.
    \begin{algorithm}[h]
        \caption{The LimeAttack algorithm}
        \label{alg:algorithm}
        \textbf{Input}: Original text $X$,target model $\mathcal{F}$\\
        \textbf{Output}: Adversarial example $X_{\text{adv}}$
        \begin{algorithmic}[1] %[1] enables line numbers
            \STATE  $X_{\text{adv}} \leftarrow X$ 
            \STATE  $set({X_\text{adv}}) \leftarrow X_{\text{adv}}$
            \STATE Compute the importance score $I(x_i)$ by LIME
      
            \STATE Sort the words with importance score $I(x_i)$
            \FOR{$i=1$ to $n$}
            \STATE Generate the candidate set $\mathcal{C}(x_i)$ 
            \ENDFOR
      
            \FOR{$X_{\text{adv}}$ in $set(X_{\text{adv}})$}
                \STATE $i$ $\leftarrow$ index of the original word
            \FOR {$c_k$ in $\mathcal{C}(x_i)$}
                \STATE $X'_{\text{adv}}$ $\leftarrow$ Replace $x_i$ with $c_k$ in  $X_{\text{adv}}$
                \STATE Add $X'_{\text{adv}}$ to the $set(X_{\text{adv}})$
            \ENDFOR
            
                \FOR{ $X'_{\text{adv}}$ in $set(X_{\text{adv}})$}
                \IF { $\mathcal{F}(X'_{\text{adv}}) \neq y_{true}$ }
                \STATE \textbf{return} $X'_{\text{adv}}$ with highest semantic similarity
                \ENDIF
                \ENDFOR
                %\Else
                %{
                \STATE $set(X_{\text{adv}})$ $\leftarrow$ Sample  $b$  adversarial examples in $set(X_{\text{adv}})$ by rules
                %}
                
            \ENDFOR
            \STATE \textbf{return} adversarial examples $X_{\text{adv}}$
        \end{algorithmic}
      \end{algorithm}

\section{Appendix I:  Qualitative Examples}
More qualify adversarial examples are listed in Table~\ref{cnn-sst}-\ref{bert-ag}

\section{Appendix J:  Limitation}
\begin{itemize}
    \item \textbf{Exploring more LLMs.} Due to limited resources, this paper only tests some popular large language models. However, there are other victim models based on other LLMs, \eg LLaMA. Hence, more victim models based on more LLMs might be studied. 
    \item \textbf{More NLP tasks.} In this paper, we only attack some classification tasks (\eg, text classification, textual entailment and zero-shot classification). It is interesting to attack other NLP applications, such as dialogue, text summarization, and machine translation. 
\end{itemize}
\section{Appendix K:  Semanticc Similarity of Different Attack Algorithms}
We have added semantic similarity in Table \ref{semantics}. Some baselines take the similarity into account during the attack, thus LimeAttack exhibits lower similarity than other methods. Considering all metrics, LimeAttack is still dominant.
\begin{table*}[htb]
    \centering
    \small
\caption{The semanticc similarity of different attack algorithms.}
\label{semantics}
\begin{tabular}{ccccccc}
    \toprule
                       &      & HLBB  & TextHoaxer & LeapAttack & TextHacker & LimeAttack \\
                       \midrule
\multirow{3}{*}{MR}    & CNN  & 97.20  & 97.11      & 97.17      & 94.56      & 95.21      \\
                       & LSTM & 97.27 & 97.27      & 97.22      & 95.01      & 95.31      \\
                       & BERT & 97.13 & 97.16      & 97.09      & 94.16      & 94.77      \\\midrule
\multirow{3}{*}{SST}   & CNN  & 97.18 & 97.22       & 97.14      & 94.02      & 94.41       \\
                       & LSTM & 97.22 & 97.21       & 97.18      & 94.58      & 94.69      \\
                       & BERT & 97.22 & 97.07      & 97.13      & 93.77      & 94.56      \\\midrule
\multirow{3}{*}{AG}    & CNN  & 97.64 & 97.62      & 97.62      & 95.71      & 96.27      \\
                       & LSTM & 97.64 & 97.58      & 97.62      & 95.46      & 96.11      \\
                       & BERT & 97.57 & 97.61      & 97.56      & 95.14      & 96.53      \\\midrule
\multirow{3}{*}{Yahoo} & CNN  & 97.75 & 97.72      & 97.71       & 95.33      & 96.21      \\
                       & LSTM & 97.71 & 97.66      & 97.67      & 95.41      & 96.41      \\
                       & BERT & 97.73 & 97.68      & 97.63      & 95.12      & 96.55     \\\bottomrule
\end{tabular}
\end{table*}

\section{Appendix L:  Comparison with Score-based Attacks}
Since LimeAttack follows the two-stage strategies samed from score-based attacks, we also take some classic score-based attacks for reference. LimeAttack and these score-based attacks have exactly the same settings. In addition, score-based attacks can obtain the probability distribution of the output, while LimeAttack does not. Therefore, we do not limit query budgets for LimeAttack and score-based attacks. As shown in Table~\ref{tab.scoreattack}, LimeAttack still achieves a higher attack success rate and semantic similarity in most cases. LimeAttack's superiority can be attributed to its focus on crucial words through the learned word importance ranking and the expanded search space with the introduction of beam search. However, LimeAttack requires more queries to compute word importance rankings because it lacks a probability distribution for the output. This situation is more obvious in long texts.
\begin{table*}[htb]
    \centering
    \caption{Comparison with other score-based attack. ASR.,\%$\uparrow$ is attack success rate, Pert.,\%$\downarrow$ is perturbation rate, Sim.,\%$\uparrow$ is semantic similarity and  Query.,$\downarrow$ is model queries.}
    \label{tab.scoreattack}
    \resizebox{\linewidth}{!}{
        \begin{tabular}{c|cccccc|c|cccccc}
            \toprule
    \textbf{Dataset}              & \textbf{Model}                 & \textbf{Attack}      & \textbf{ASR.$\uparrow$} & \textbf{Pert.$\downarrow$} & \textbf{Sim.$\uparrow$}  & \textbf{Query.$\downarrow$} & \textbf{Dataset}                & \textbf{Model}                 & \textbf{Attack}     & \textbf{ASR.$\uparrow$} & \textbf{Pert.$\downarrow$} & \textbf{Sim.$\uparrow$}  & \textbf{Query.$\downarrow$} \\\midrule
    \multirow{10}{*}{MR} & \multirow{4}{*}{CNN}  & TF          & 60.9  & 5.88  & 94.21 & 51.84  & \multirow{6}{*}{AG}    & \multirow{3}{*}{CNN}  & TF         & 32.1  & 5.96  & 94.65 & \textbf{43.67}  \\
    &                       & PWWS        & 62.4  & 5.88  & 92.34 & 144.37 &                        &                       & PWWS       & 32.1  & 5.94  & 94.85 & 47.68  \\
    &                       & Bert-Attack & 46.3  & 5.75  & 94.51 & \textbf{28.25}  &                        &                       & \textbf{LimeAttack} & \textbf{38.1}  & \textbf{4.55}  & \textbf{96.53} & 879.23 \\\cline{9-14}
    &                       & \textbf{LimeAttack}  & \textbf{62.5}  & \textbf{5.60}  & \textbf{95.33} & 268.94 &                        & \multirow{3}{*}{LSTM} & TF         & 30.5  & 5.51  & 95.40 & \textbf{46.93}  \\\cline{2-7}
    & \multirow{3}{*}{LSTM} & TF          & \textbf{65.8}  & 5.63  & 94.56 & 49.96  &                        &                       & PWWS       & 32.1  & 5.94  & 94.85 & 47.68  \\
    &                       & Bert-Attack & 50.2  & 5.77  & 94.4  & \textbf{28.53}  &                        &                       & \textbf{LimeAttack} & \textbf{35.4}  & \textbf{4.55}  & \textbf{96.13} & 975.35 \\\cline{8-14}
    &                       & \textbf{Limeattack}  & 61.2  & \textbf{5.51}  & \textbf{95.44} & 253.07 & \multirow{4}{*}{SST-2} & \multirow{2}{*}{CNN}  & TF         & \textbf{51.0}  & \textbf{5.96}  & 93.83 & \textbf{51.67}  \\\cline{2-7}
    & \multirow{3}{*}{BERT} & TF          & 46.5  & 5.68  & 94.43 & 51.48  &                        &                       & \textbf{LimeAttack} & \textbf{51.0}  & 5.99  & \textbf{94.90} & 150.08 \\\cline{9-14}
    &                       & Bert-Attack & 35.0  & 5.82  & 94.64 & \textbf{28.59}  &                        & \multirow{2}{*}{LSTM} & TF         & \textbf{52.1}  & \textbf{5.93}  & 93.54 & \textbf{50.7}   \\
    &                       & \textbf{LimeAttack}  & \textbf{47.6}  & \textbf{5.59}  & \textbf{94.99} & 821.28 &                        &                       & \textbf{LimeAttack} & 50.5  & 6.13  & \textbf{94.70} & 320.45\\\bottomrule
    \end{tabular}}
    \end{table*}
\section{Appendix M:  Evaluation on Defense Methods}
We used A2T (The core part of A2T is a new and cheaper word substitution attack optimized for adversarial training) and ASCC  to enhance the defense ability of BERT on MR and SST datasets, and conducted attack experiments on this basis. As shown in Table~\ref{defense}. Even after adversarial training and enhancement, our algorithm still has a certain attack effect on these defense methods. Compared with A2T, ASCC has better defense effect and improves a certain degree of model robustness.
\begin{table*}[htb]
\centering
\caption{The attack performance of different attack algorithms on A2T and ASCC defense methods and original target models in BERT-MR and BERT-SST.}
\label{defense}
\begin{tabular}{ccccccccccccc}
 \toprule
\multirow{2}{*}{} & \multicolumn{2}{c}{origin BERT-MR} & \multicolumn{2}{c}{A2T } & \multicolumn{2}{c}{ASCC } & \multicolumn{2}{c}{origin BERT-SST} & \multicolumn{2}{c}{A2T } & \multicolumn{2}{c}{ASCC} \\
                  & ASR              & PERT            & ASR            & PERT           & ASR             & PERT           & ASR              & PERT             & ASR             & PERT           & ASR             & PERT            \\\midrule
HLBB              & 26.6             & 5.6             & 23.5           & 5.6            & 20.1            & 5.6            & 23.0               & 5.8              & 21.3            & 6.0              & 19.3            & 6.1             \\
TextHoaxer        & 27.0               & 5.5             & 24.3           & 5.6            & 21.2            & 5.7            & 24.9             & 5.8              & 21.8            & 5.9            & 20.1            & 5.9             \\
LeapAttack        & 26.5             & 5.4             & 24.0             & 5.6            & 22.3            & 5.6            & 26.1             & 5.8              & 21.7            & 5.9            & 19.6            & 6.1             \\
TextHoaxer        & 26.5             & 6.5             & 24.1           & 6.6            & 22.5            & 6.6            & 25.4             & 6.3              & 22.1            & 6.3            & 19.1            & 6.6             \\
LimeAttack        & 29.2             & 5.9             & 25.7           & 5.8            & 23.4            & 5.8            & 27.8             & 5.7              & 22.7            & 5.9            & 20.3            & 6.1    \\\bottomrule        
\end{tabular}
\end{table*}

\section{Appendix N:  Convergence of Attack Performance}
\subsection{convergence of attack success rate}
We have conduct further evaluations on defense methods to validate their effectiveness. As shown in Table~\ref{tab.convergenceattack}, LimeAttack achieves better attack success rate than other attacks. Attack success rate without considering the query budget is more of an ideal situation. It shows the upper limit of an attack algorithm. High query budget is equivalent to traverse the solution space and will approximate the asr and pert upper limit of victim model; However, asr and pert will interact with each other, resulting in the upper limit of asr and pert not being in the same direction. Therefore, for some victim models (LSTM-AG and BERT-Yahoo), limeattack's pert is the lowest, but not the optimal asr (very close).
\begin{table*}[htb]
  \caption{Different attack algorithms on different model and datasets under query is 1000.}
  \label{tab.convergenceattack}
  \resizebox{\linewidth}{!}{
  \begin{tabular}{ccccccccccccccc}
    \toprule
  \multirow{2}{*}{} & \multicolumn{2}{c}{CNN\_MR}  & \multicolumn{2}{c}{CNN\_SST} & \multicolumn{2}{c}{LSTM\_MR} & \multicolumn{2}{c}{LSTM\_SST} & \multicolumn{2}{c}{LSTM\_AG} & \multicolumn{2}{c}{BERT\_SST} & \multicolumn{2}{c}{BERT\_Yahoo} \\
                    & ASR           & PERT         & ASR           & PERT         & ASR           & PERT         & ASR            & PERT         & ASR           & PERT         & ASR            & PERT         & ASR             & PERT          \\\midrule
  HLBB              & 55.6          & 5.6          & 43.4          & 6.4          & 54.5          & 5.6          & 43.3           & 6.4          & 30.4          & 5.5          & 30.3           & 6.7          & 62.2            & 6.7           \\
  TextHoaxer        & 55.6          & \textbf{5.4} & 43.9          & 6.4          & 52.9          & \textbf{5.4} & \textbf{45.5}  & 6.3          & 31.1          & 5.8          & 35.9           & 6.6          & 63.2            & 6.6           \\
  LeapAttack        & 56.4          & 5.5          & 44.3          & 6.5          & 54.6          & 5.5          & 44.3           & 6.2          & 31.3          & \textbf{5.3} & 37.5           & 6.2          & 63.1            & 6.4           \\
  TextHoaxer        & 59.2          & 5.6          & 38.0          & 6.7          & 56.0          & 5.6          & 44.0           & 6.5          & \textbf{32.0} & 5.8          & 38.0           & \textbf{6.0} & \textbf{67.2}   & 6.4           \\
  LimeAttack        & \textbf{59.4} & 5.7          & \textbf{48.6} & \textbf{6.0} & \textbf{59.3} & 5.5          & \textbf{45.5}  & \textbf{5.9} & 31.2          & \textbf{5.3} & \textbf{42.5}  & 6.1          & 66.0            & \textbf{6.2} \\\bottomrule
  \end{tabular}}
  \end{table*}
\subsection{convergence of perturbation rate}
\begin{figure}[h]
    \centering
    \includegraphics[scale=0.2]{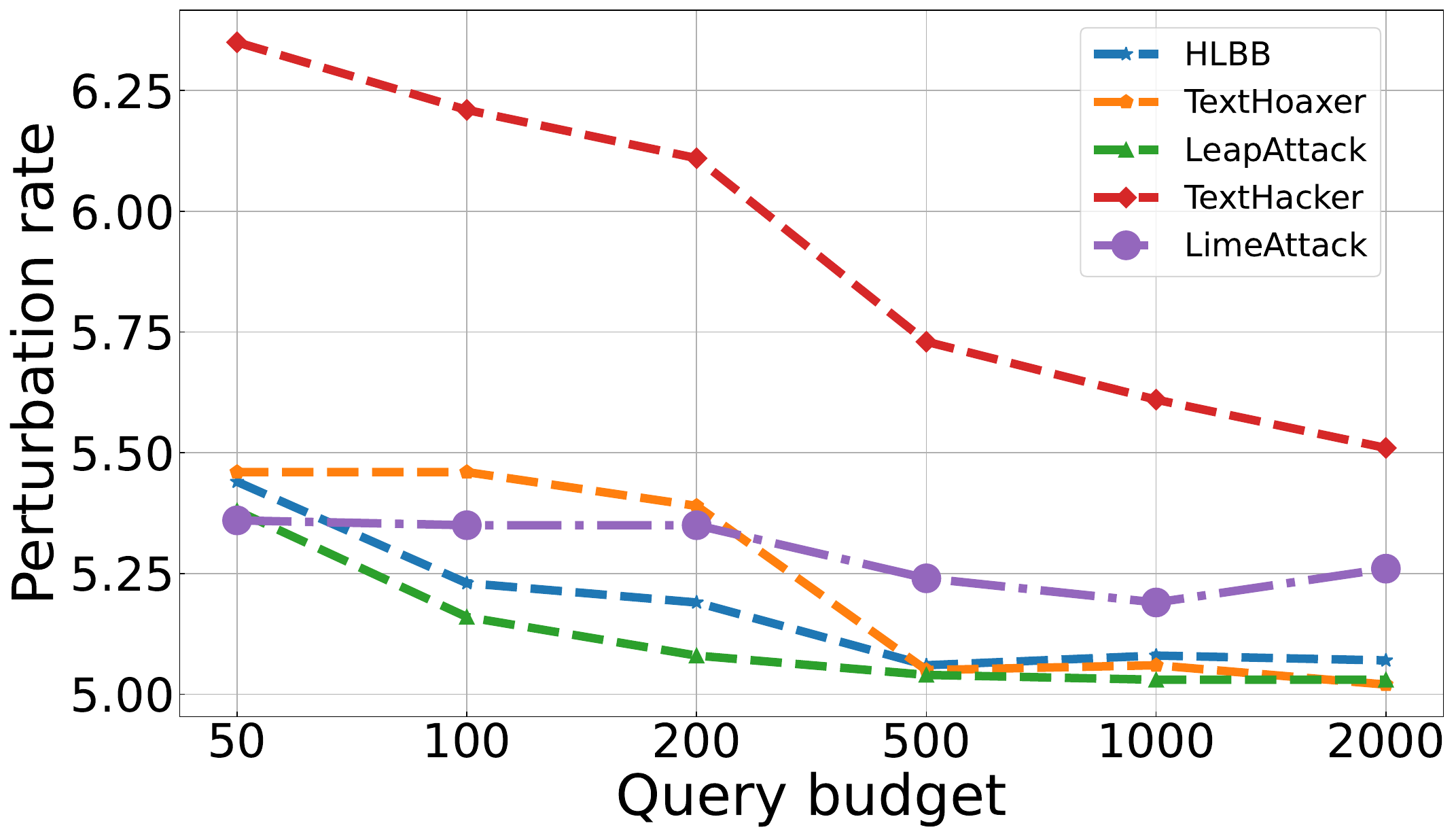}%mr_bert.pdf}
    \caption{Perturbation rate of different  attacks on CNN-MR.}
    \label{fig.query}
    \end{figure}

We list convergence behavior of different attack. As shown in the figure~\ref{fig.query}. Due to the use of complex optimization algorithms in previous algorithms, it does require a large number of queries to complete this part of optimization; Therefore, previous algorithms often have a good perturbation rates.

\section{Appendix O:  Comparison with SHAP and Non-linear Models}
In a hard-label setting, model's logits are unavailable and model query budget is tiny. We list the result of attack success rate of different word importance ranking calculation under different query budgets. As shown in the Table~\ref{shap}, compared to LIME, attack success rate and perturbation rate of SHAP or non-linear models do not have significant advantages in tiny query budgets. Considering the time complexity, we adopt LIME to calculate word importance ranking in the main text.

\begin{table}[H]
\centering
\caption{Evaluation of different word importance ranking calculation on CNN-MR and BERT-SST under different query budgets.}
    \label{shap}
    \resizebox{\linewidth}{!}{
\begin{tabular}{ccccccccc}
\toprule
\multirow{3}{*}{} & \multicolumn{4}{c}{query budgets 100}                     & \multicolumn{4}{c}{query budgets 2000}                    \\
                  & \multicolumn{2}{c}{CNN-MR} & \multicolumn{2}{c}{BERT-SST} & \multicolumn{2}{c}{CNN-MR} & \multicolumn{2}{c}{BERT-SST} \\
                  & ASR          & PERT        & ASR           & PERT         & ASR          & PERT        & ASR           & PERT         \\ \midrule
LIME              & 49.9         & 5.3         & 27.8          & 5.7          & 59.4         & 5.7         & 42.5          & 6.1          \\
SHAP              & 49.7         & 5.2         & 27.7          & 5.7          & 61.2         & 5.8         & 44.3          & 6.3          \\
Decision Tree     & 50.1         & 5.3         & 27.9          & 5.8          & 61.6         & 5.8         & 44.1          & 6.4   \\\bottomrule      
\end{tabular}}
\end{table}

%\bibliography{ref}
%\bibliographystyle{unsrt} 
\begin{table*}[htb]
  \centering
  \caption{The adversarial example crafted by different attack algorithms on  CNN using SST-2 dataset.  Replacement words are represented in  \textcolor{red}{red}. Query.$\downarrow$ is model query numbers.}
  \label{cnn-sst}
  \resizebox{\linewidth}{!}{
  \begin{tabular}{ccc}
  \toprule
  \textbf{Attack} & \textbf{Texts} & \textbf{Query.} \\
  \midrule
  No Attack & \multicolumn{1}{m{11.6cm}}{It allows us hope that nolan is poised to embark a major career as a commercial yet inventive filmmaker.
  }& \makecell{0} \\\midrule
  HLBB & \multicolumn{1}{m{11.6cm}}{It allows us hope that nolan is poised to \red{incur} a major career as a commercial yet \red{ingenuity} filmmaker.
  }& \makecell{2062} \\\midrule
  TextHoaxer & \multicolumn{1}{m{11.6cm}}{It allows us hope that nolan is poised to \red{start} a major career as a commercial yet \red{contrivance} filmmaker.
  }& \makecell{48} \\\midrule
  LeapAttack & \multicolumn{1}{m{11.6cm}}{It allows us hope that nolan is poised to embark a major career as a commercial yet \red{contrivance} filmmaker.
  }& \makecell{30} \\\midrule
  TextHacker & \multicolumn{1}{m{11.6cm}}{It allows us hope that nolan is \red{readies} to embark a major career as a commercial yet \red{creative} filmmaker.}& \makecell{101} \\\midrule
  \textbf{LimeAttack}  & \multicolumn{1}{m{11.6cm}}{It allows us hope that nolan is poised to embark a major career as a commercial yet \red{contrivance} filmmaker.}& \makecell{{43}} \\
  \bottomrule
  \end{tabular}}
  \end{table*}

\begin{table*}[htb]

\centering
\caption{The adversarial example crafted by different attack algorithms on   BERT using SST-2 dataset.  Replacement words are represented in  \textcolor{red}{red}. Query.$\downarrow$ is model query numbers.}
\label{bert-sst}
\resizebox{\linewidth}{!}{
\begin{tabular}{ccc}
\toprule
\textbf{Attack} & \textbf{Texts} & \textbf{Query.} \\
\midrule
No Attack & \multicolumn{1}{m{11.6cm}}{The   acting,costumes,music,cinematogrtaphy and sound are all astounding given the   production's austere locales.}& \makecell{0} \\\midrule
HLBB & \multicolumn{1}{m{11.6cm}}{The   acting,costumes,music,cinematogrtaphy and sound are all \red{stupendous} given the   production's austere locales.}& \makecell{35} \\\midrule
TextHoaxer & \multicolumn{1}{m{11.6cm}}{The   acting,costumes,music,cinematogrtaphy and sound are all \red{staggering} given the   production's austere locales.}& \makecell{45} \\\midrule
LeapAttack & \multicolumn{1}{m{11.6cm}}{the   acting,costumes,music,cinematogrtaphy and sound are all astounding   \red{dispensed}  the production's austere   locales.}& \makecell{35} \\\midrule
TextHacker & \multicolumn{1}{m{11.6cm}}{the   \red{provisonal},costumes,music,cinematogrtaphy and sound \red{sunt} all \red{startling} given   the production's \red{stoic} locales.}& \makecell{101} \\\midrule

\textbf{LimeAttack}  & \multicolumn{1}{m{11.6cm}}{the   acting,costumes,music,cinematogrtaphy and sound are all \red{staggering} given the   production's austere locales.}& \makecell{{25}} \\
\bottomrule
\end{tabular}}

\end{table*}

%\bibliographystyle{named}
%\bibliography{ijcai23}

\begin{table*}[htb]
  \centering
  \caption{The adversarial example crafted by different attack algorithms on   LSTM using Yahoo dataset.  Replacement words are represented in  \textcolor{red}{red}. Query.$\downarrow$ is model query numbers.}
  \label{lstm-yahoo}
  \resizebox{\linewidth}{!}{
  \begin{tabular}{ccc}
  \toprule
  \textbf{Attack} & \textbf{Texts} & \textbf{Query.} \\
  \midrule
  No Attack & \multicolumn{1}{m{11.6cm}}{In basketball whats a suicide? is it like running back and forth? its an exercise where you run the entire court touching down in intnervals until youve completed the exercise on both sides of the court.
  }& \makecell{0} \\\midrule
  HLBB & \multicolumn{1}{m{11.6cm}}{In \red{basket} whats a suicide? is it like running back and forth? its an exercise where you run the entire court touching down in intnervals until youve completed the exercise on both sides of the court.
  }& \makecell{6} \\\midrule
  TextHoaxer & \multicolumn{1}{m{11.6cm}}{In \red{wildcats} whats a suicide? is it like running back and forth? its an exercise where you run the entire court touching down in intnervals until youve completed the exercise on both sides of the court
  }& \makecell{6} \\\midrule
  LeapAttack & \multicolumn{1}{m{11.6cm}}{In \red{wildcats} whats a suicide? is it like running back and forth? its an exercise where you run the entire court touching down in intnervals until youve completed the exercise on both sides of the court.
  }& \makecell{6} \\\midrule
  TextHacker & \multicolumn{1}{m{11.6cm}}{In basketball whats a suicide? is it like running back and forth? its an exercise where you run the entire court touching down in intnervals until \red{havent} completed the exercise on both sides of the court.}& \makecell{101} \\\midrule

  \textbf{LimeAttack}  & \multicolumn{1}{m{11.6cm}}{In \red{basketballs} whats a suicide? is it like running back and forth? its an exercise where you run the entire court touching down in intnervals until youve completed the exercise on both sides of the court.}& \makecell{{39}} \\
  \bottomrule
  \end{tabular}}
  \end{table*}

\begin{table*}[htb]
\centering
\caption{The adversarial example crafted by different attack algorithms on  CNN using Yahoo dataset.  Replacement words are represented in  \textcolor{red}{red}. Query.$\downarrow$ is model query numbers.}
\label{cnn-yahoo}
\resizebox{\linewidth}{!}{
\begin{tabular}{ccc}
\toprule
\textbf{Attack} & \textbf{Texts} & \textbf{Query.} \\
\midrule
No Attack & \multicolumn{1}{m{11.6cm}}{Who was the first indian who became the member of english parliament? dadabhai naoroji preeminent pioneer of indian nationalism freedom fighter and educationist the first indian to become member of british parliament 1862 congress president thrice the grand old man of india.
}& \makecell{0} \\\midrule
HLBB & \multicolumn{1}{m{11.6cm}}{Who was the first indian who became the member of english parliament? dadabhai naoroji preeminent \red{groundbreaking} of indian nationalistic freedom \red{hunter} and educationist the first indian to become member of british parliament 1862 congress president thrice the grand old man of india.
}& \makecell{116} \\\midrule
TextHoaxer & \multicolumn{1}{m{11.6cm}}{Who was the first indian who became the member of english parliament? dadabhai naoroji preeminent pioneer of indian nationalism freedom fighter and educationist the first indian to become member of british parliament 1862 congress president thrice the \red{immense} old man of \red{indian}.
}& \makecell{440} \\\midrule
LeapAttack & \multicolumn{1}{m{11.6cm}}{Who was the first indian who became the member of english parliament? dadabhai naoroji preeminent pioneer of indian nationalism \red{liberty} \red{hunters} and educationist the first indian to become member of british parliament 1862 congress president thrice the grand old man of india.
}& \makecell{1411} \\\midrule
TextHacker & \multicolumn{1}{m{11.6cm}}{Who was the first indian who became the member of english parliament? dadabhai naoroji preeminent \red{pioneers} of indian nationalism freedom fighter and educationist the first indian to become member of british \red{chambre} 1862 congress president thrice the grand old man of india.}& \makecell{101} \\\midrule

\textbf{LimeAttack}  & \multicolumn{1}{m{11.6cm}}{Who was the first indian who became the member of english parliament? dadabhai naoroji preeminent pioneer of indian nationalism freedom fighter and educationist the first indian to become member of british \red{legislature} 1862 congress president thrice the grand old man of india.}& \makecell{{45}} \\
\bottomrule
\end{tabular}}
\end{table*}

\begin{table*}[htb]

\centering
\caption{The adversarial example crafted by different attack algorithms on  CNN using MR dataset.  Replacement words are represented in  \textcolor{red}{red}. Query.$\downarrow$ is model query numbers.}
\label{cnn-mr}
\resizebox{\linewidth}{!}{
\begin{tabular}{ccc}
\toprule
\textbf{Attack} & \textbf{Texts} & \textbf{Query.} \\
\midrule
No Attack & \multicolumn{1}{m{11.6cm}}{Those outside show business will enjoy a close look at people they do n't really want to know.
}& \makecell{0} \\\midrule
HLBB & \multicolumn{1}{m{11.6cm}}{Those outside show business will enjoy a \red{nearby} look at people they do n't really want to know.
}& \makecell{2241} \\\midrule
TextHoaxer & \multicolumn{1}{m{11.6cm}}{Those outside show business will \red{recieve} a close look at people they do n't really want to know.
}& \makecell{202} \\\midrule
LeapAttack & \multicolumn{1}{m{11.6cm}}{Those outside show business will \red{like} a close glanced at people they do n't really want to know.
}& \makecell{1431} \\\midrule
TextHacker & \multicolumn{1}{m{11.6cm}}{Those outside show \red{companies} will \red{experience} a close \red{glance} at \red{volk} they do n't really want to know.}& \makecell{103} \\\midrule
\textbf{LimeAttack}  & \multicolumn{1}{m{11.6cm}}{Those outside show business will \red{recieve} a close glanced at people they do n't really want to know}& \makecell{{53}} \\
\bottomrule
\end{tabular}}
\end{table*}

\begin{table*}[htb]

\centering
\caption{The adversarial example crafted by different attack algorithms on  LSTM using MR dataset.  Replacement words are represented in  \textcolor{red}{red}. Query.$\downarrow$ is model query numbers.}
\label{lstm-mr}
\resizebox{\linewidth}{!}{
\begin{tabular}{ccc}
\toprule
\textbf{Attack} & \textbf{Texts} & \textbf{Query.} \\
\midrule
No Attack & \multicolumn{1}{m{11.6cm}}{I'm convinced i could keep a family of five blind , crippled , amish people alive in this situation better than these british soldiers do at keeping themselves kicking.
}& \makecell{0} \\\midrule
HLBB & \multicolumn{1}{m{11.6cm}}{I'm convinced i could keep a family of five blind , \red{invalids} , amish people alive in this situation better than these british soldiers do at keeping themselves kicking.
}& \makecell{2110} \\\midrule
TextHoaxer & \multicolumn{1}{m{11.6cm}}{I'm \red{gratified} i could keep a family of five blind , crippled , amish people alive in this situation better than these british soldiers do at keeping themselves kicking.
}& \makecell{219} \\\midrule
LeapAttack & \multicolumn{1}{m{11.6cm}}{I'm \red{contented} i could keep a family of five blind , \red{paralytic}, amish people alive in this \red{plight} better than these british soldiers do at keeping themselves kicking.
}& \makecell{2162} \\\midrule
TextHacker & \multicolumn{1}{m{11.6cm}}{I'm convinced i could keep a family of five blind , \red{handicapped} , amish people \red{lively} in this situation better than these british soldiers do at keeping themselves kicking.}& \makecell{101} \\\midrule
\textbf{LimeAttack}  & \multicolumn{1}{m{11.6cm}}{I'm \red{gratified} i could keep a family of five blind , crippled , amish people alive in this situation better than these british soldiers do at keeping themselves kicking.}& \makecell{{50}} \\
\bottomrule
\end{tabular}}
\end{table*}

\begin{table*}[htb]

\centering
\caption{The adversarial example crafted by different attack algorithms on  LSTM using AG dataset.  Replacement words are represented in  \textcolor{red}{red}. Query.$\downarrow$ is model query numbers.}
\label{lstm-ag}
\resizebox{\linewidth}{!}{
\begin{tabular}{ccc}
\toprule
\textbf{Attack} & \textbf{Texts} & \textbf{Query.} \\
\midrule
No Attack & \multicolumn{1}{m{11.6cm}}{Spaniards to run luton airport after 551 m deal luton , cardiff and belfast international airports are to fall into the hands of a spanish toll motorways operator through a 551 m takeover of the aviation group tbi by a barcelona based abertis infrastructure.
}& \makecell{0} \\\midrule
HLBB & \multicolumn{1}{m{11.6cm}}{spaniards to \red{executes} luton airport after 551 m deal luton , cardiff and belfast international airports are to fall into the hands of a spanish toll motorways operator through a 551 m takeover of the \red{aeroplanes} group tbi by a barcelona based abertis infrastructure.
}& \makecell{969} \\\midrule
TextHoaxer & \multicolumn{1}{m{11.6cm}}{Spaniards to run luton airport after 551 m deal luton , cardiff and belfast international airports are to fall into the \red{manaus} of a spanish toll motorways exploiter through a 551 m \red{coup} of the aviation group tbi by a barcelona based abertis infrastructure.
}& \makecell{727} \\\midrule
LeapAttack & \multicolumn{1}{m{11.6cm}}{Spaniards to run luton airport after 551 m deal luton , cardiff and belfast international airports are to fall into the hands of a spanish toll motorways operator through a 551 m takeover of the \red{aeroplanes} group tbi by a barcelona based abertis infrastructure.
}& \makecell{2148} \\\midrule
TextHacker & \multicolumn{1}{m{11.6cm}}{Spaniards to \red{implementing} luton airport after 551 m deal luton , cardiff and belfast international airports \red{represent} to fall into the hands of a spanish toll motorways operator through a 551 m takeover of the aviation group tbi by a barcelona based abertis infrastructure.}& \makecell{101} \\\midrule
\textbf{LimeAttack}  & \multicolumn{1}{m{11.6cm}}{Spaniards to run luton \red{luton} after 551 m deal luton , cardiff and belfast international airports are to fall into the hands of a spanish toll motorways operator through a 551 m takeover of the aviation group tbi by a barcelona based abertis infrastructure.}& \makecell{{92}} \\
\bottomrule
\end{tabular}}
\end{table*}

\begin{table*}[htb]

\centering
\caption{The adversarial example crafted by different attack algorithms on  CNN using AG dataset.  Replacement words are represented in  \textcolor{red}{red}. Query.$\downarrow$ is model query numbers.}
\label{cnn-ag}
\resizebox{\linewidth}{!}{
\begin{tabular}{ccc}
\toprule
\textbf{Attack} & \textbf{Texts} & \textbf{Query.} \\
\midrule
No Attack & \multicolumn{1}{m{11.6cm}}{Eisner says ovitz required oversight daily michael d eisner appeared for a second day of testimony in the shareholder lawsuit over the lucrative severance package granted to michael s ovitz.
}& \makecell{0} \\\midrule
HLBB & \multicolumn{1}{m{11.6cm}}{Eisner says ovitz required oversight daily michael d eisner appeared for a second \red{weekly} of testimony in the shareholder lawsuit over the lucrative severance package granted to michael s ovitz.
}& \makecell{31} \\\midrule
TextHoaxer & \multicolumn{1}{m{11.6cm}}{Eisner says ovitz required oversight daily michael d eisner appeared for a second day of testimony in the shareholder lawsuit over the \red{interesting} severance package granted to michael s ovitz.
}& \makecell{48} \\\midrule
LeapAttack & \multicolumn{1}{m{11.6cm}}{Eisner says ovitz \red{needing} oversight daily michael d eisner appeared for a second day of testimony in the shareholder lawsuit over the lucrative severance package granted to michael s ovitz.
}& \makecell{14} \\\midrule
TextHacker & \multicolumn{1}{m{11.6cm}}{Eisner says ovitz required \red{surveillance} \red{everyday} michael d eisner appeared for a second day of \red{testimonies} in the shareholder lawsuit over the \red{rewarding} severance package granted to michael s ovitz.}& \makecell{101} \\\midrule
\textbf{LimeAttack}  & \multicolumn{1}{m{11.6cm}}{Eisner says ovitz required oversight daily michael d eisner appeared for a second day of testimony in the \red{proprietors} lawsuit over the lucrative severance package granted to michael s ovitz.}& \makecell{{34}} \\
\bottomrule
\end{tabular}}
\end{table*}
\begin{table*}[htb]
  \centering
  \caption{The adversarial example crafted by different attack algorithms on   BERT using AG dataset.  Replacement words are represented in  \textcolor{red}{red}. Query.$\downarrow$ is model query numbers.}
  \label{bert-ag}
  \resizebox{\linewidth}{!}{
  \begin{tabular}{ccc}
  \toprule
  \textbf{Attack} & \textbf{Texts} & \textbf{Query.} \\
  \midrule
  No Attack & \multicolumn{1}{m{11.6cm}}{Cray promotes two execs ly huong pham becomes the supercomputer maker's senior vice presdent of operations,and peter ungaro is made senior vice president for sales,marketing and services.}& \makecell{0} \\\midrule
  HLBB & \multicolumn{1}{m{11.6cm}}{Cray promotes two execs ly huong pham \red{buys} the supercomputer maker's senior vice presdent of operations,and peter ungaro is made senior \red{obscene} \red{chairperson} for sales,marketing and services.  }& \makecell{3811} \\\midrule
  TextHoaxer & \multicolumn{1}{m{11.6cm}}{\red{Hucknall} promotes two execs ly huong pham becomes the supercomputer maker's senior vice president of \red{surgical}, and peter ungaro is made senior vice president for sales, marketing and services.
  }& \makecell{94} \\\midrule
  LeapAttack & \multicolumn{1}{m{11.6cm}}{Cray promotes two execs ly huong pham becomes the \red{quadrillion} maker's senior vice presdent of operations,and peter ungaro is made senior vice president for  sales,marketing and services.    }& \makecell{42} \\\midrule
  TextHacker & \multicolumn{1}{m{11.6cm}}{Cray promotes two \red{ceos} ly huong pham becomes the supercomputer maker's senior \red{prostitution} presdent of operations,and peter ungaro is made senior vice president for \red{selling},marketing and services.}& \makecell{101} \\\midrule

  \textbf{LimeAttack}  & \multicolumn{1}{m{11.6cm}}{Cray promotes two execs ly huong pham becomes the \red{thermonuclear} maker's senior vice presdent of operations,and peter ungaro is made senior vice president for sales,marketing and services.}& \makecell{{39}} \\
  \bottomrule
  \end{tabular}}
  \end{table*}

\end{document}